\documentclass[11pt,letterpaper]{article}

\usepackage[margin=1in]{geometry}
\usepackage[T1]{fontenc}
\usepackage{newtxtext,newtxmath}
\usepackage{microtype}

\usepackage{graphicx}
\usepackage[figuresright]{rotating}
\usepackage{booktabs}
\usepackage{makecell}
\usepackage{tabularx}
\usepackage{subcaption}

\usepackage{authblk}
\usepackage[numbers,sort&compress]{natbib}

\usepackage[
    bookmarks=false,
    colorlinks=true,
    linkcolor=blue,
    citecolor=blue,
    urlcolor=blue
]{hyperref}

\title{Physics-Informed Feature Representations
and Pretrained Tabular Foundation Models
for Generalizable Melt Pool Geometry Prediction}

\author[1]{Rafid Enayet}
\author[2]{Imtiaz Ahmed}
\author[1]{Md Habibor Rahman\thanks{
Corresponding author:
\href{mailto:habiborrahman@sc.edu}
{habiborrahman@sc.edu}}}

\affil[1]{\small {University of South Carolina,
Columbia, SC 29208, USA}}

\affil[2]{\small {West Virginia University,
Morgantown, WV 26506, USA}}

\date{}

\begin{document}

\maketitle

\begin{abstract}
Developing generalizable, data-efficient machine learning (ML) models for advanced manufacturing processes remains a fundamental challenge in smart manufacturing systems, particularly when experimental datasets are small and process-property relationships must transfer across materials and operating conditions. Laser powder bed fusion (LPBF) exemplifies this challenge, where build quality is highly sensitive to the coupled effects of laser processing conditions and thermophysical properties of the material, and the melt pool geometry – encompassing width and depth – together with the prevailing melting mode (conduction, transition, or keyhole) are critical determinants of process stability, defect formation, and part integrity. Conventional data-driven models predict these responses directly from dimensional process parameters, sometimes augmented by material properties. However, such representations encode laser-material interaction physics only implicitly, limiting their ability to generalize across materials or process conditions without retraining. This study investigates whether physics-informed dimensionless parameters – specifically normalized enthalpy and Peclet number, derived by combining laser processing conditions with material thermophysical properties – provide a more transferable and effective feature representation than raw process parameters for data-driven melt pool modeling. A systematic benchmark is conducted across three material systems (316L stainless steel, IN625, and IN718 nickel-based alloys) and their combination, evaluating seven models spanning classical ML algorithms – Gaussian Process, Support Vector Machine, Random Forest, and XGBoost – and three pretrained Tabular Foundation Models – TabPFN, TabDPT, and TabICLv2 – for both regression of melt pool depth and width and classification of melting modes. The results demonstrate that dimensionless representations improve predictive performance across most evaluated conditions, with the most substantial improvement observed in cross-material melt pool depth prediction, where the model achieved an $R^2$ of 0.96. Moreover, the study finds that pretrained tabular foundation models consistently outperform classical ML baselines under both feature representations, demonstrating that synthetic pretraining priors transfer effectively to the structured, physics-governed relationships of LPBF. These findings establish that combining physically meaningful dimensionless representations with modern pretrained tabular models offers a principled strategy for building generalizable process-property models under limited experimental data, with direct implications for accelerating process-window identification and reducing reliance on material-specific trial-and-error experimentation in laser-based manufacturing systems.
\end{abstract}

\vspace{0.5em}

\section{Introduction}
\label{Introduction}

Integrating machine learning (ML) into advanced manufacturing systems has emerged as a transformative approach to accelerate product design and process development, enable real-time quality control, and reduce reliance on costly trial-and-error experimentation \cite{Leng2026, Wang2018, Sharp2018}. Within this paradigm, a persistent challenge is\enlargethispage{-3\baselineskip} building models that generalize reliably beyond the specific material system and operating conditions on which they were trained \cite{Wang2022a, Ding2024}. This is a major concern in industrial practice: data-driven process-property models calibrated for one alloy must typically be retrained from scratch when the material changes, an expensive requirement given the cost and time associated with manufacturing experiments \cite{Vendrell-Gallart2025}. The root cause of this fragility lies not in data scarcity alone -- though that is endemic to advanced manufacturing processes -- but also in the choice of input representation. Process parameters such as laser power, scan speed, and beam diameter carry implicit dependencies on material thermophysical properties that the model must reconstruct from limited observations, often failing to generalize across material systems \cite{Inam2024, Mosallanejad2024}. Addressing this challenge in a principled way requires both rethinking how domain physics are encoded into the feature space and evaluating whether emerging data-efficient modeling architectures can provide the inductive biases needed to offset the structural disadvantage of small, material-specific datasets.

\begin{figure*}[!htb] 
    \centering
    
    \begin{minipage}[b]{0.22\textwidth}
        \centering
        \includegraphics[width=\linewidth, height=3.5cm]{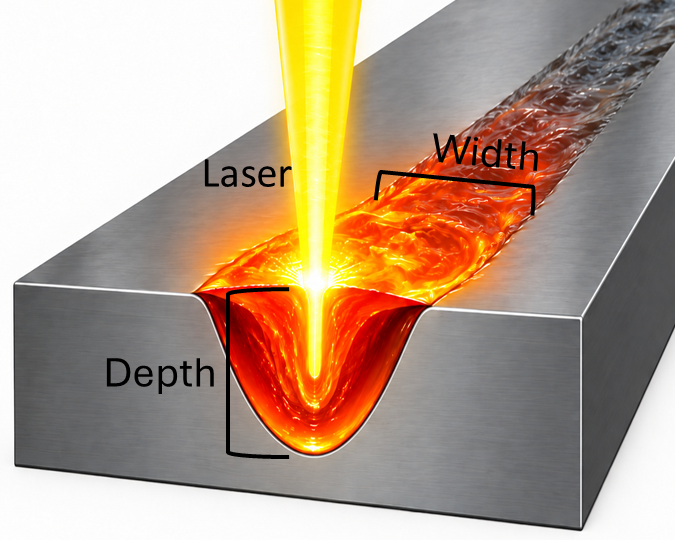}
        \subcaption{}
        \label{fig:fig1}
    \end{minipage}
    \hfill
    \begin{minipage}[b]{0.22\textwidth}
        \centering
        \includegraphics[width=\linewidth, height=3.5cm]{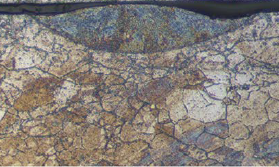}
        \subcaption{}
        \label{fig:fig2}
    \end{minipage}
    \hfill      
    \begin{minipage}[b]{0.22\linewidth}
        \centering
        \includegraphics[width=\linewidth, height=3.5cm]{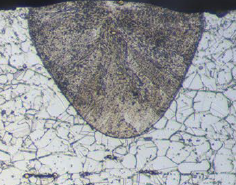}
        \subcaption{}
         \label{fig:fig3}
    \end{minipage}
    \hfill
    \begin{minipage}[b]{0.22\linewidth}
        \centering
        \includegraphics[width=\linewidth, height=3.5cm]{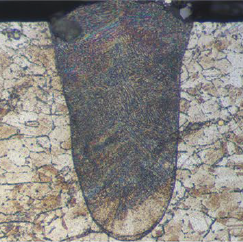}
        \subcaption{}
        \label{fig:fig4}
    \end{minipage}
    
    \caption{(a) Schematic representation of melt pool track in LPBF; (b), (c), (d) are representative micrographs of melt pool cross sections ranging from conduction to keyholing mode \cite{WeaverJordanS.DeisenrothDavidMekhontsevSergeyHeigelJarredLane2023}.}
    \label{fig:meltpool_overview}
\end{figure*}

Laser powder bed fusion (LPBF) is one of the most industrially significant and methodologically demanding contexts for examining these questions. As an established metal additive manufacturing (AM) process that can produce geometrically complex, near-net-shape components from a broad range of engineering alloys, LPBF has attracted considerable attention for aerospace, biomedical, and energy applications \cite{bushra2026reverse,shahriar2026knowledge}. The quality and structural integrity of LPBF-built parts depend critically on the melt pool geometry formed during laser scanning, as shown in Figure \ref{fig:meltpool_overview}. The melt pool width and depth govern inter-layer bonding, determine the onset of defect modes -- lack-of-fusion porosity, balling, and keyhole collapse -- and establish the thermal gradients responsible for microstructure evolution \cite{King2014, Wang2022}. Beyond geometry, identifying the prevailing melting mode (conduction, transition, or keyhole) is essential for defining stable processing windows, since each regime is associated with distinct defect mechanisms and part-quality outcomes \cite{Hofmann2026, hasan2023process}. Despite this importance, systematically characterizing melt pool behavior experimentally across the full process parameter space is prohibitively time-intensive and requires extensive single-track campaigns for each new material. \cite{Schlenger2025}. The combination of small experimental datasets, complex multi-physics interactions, and the need to transfer knowledge across materials makes LPBF an ideal and demanding benchmark for evaluating generalization strategies in manufacturing ML systems.
 
Recognizing the limitations of purely physics-based analytical models, researchers have applied a broad range of ML methods to melt pool geometry prediction in LPBF \cite{Naderi2023}. Early efforts employed Gaussian Process (GP), Support Vector Machine (SVM), Random Forest (RF), and multivariate adaptive regression splines using laser power, scan speed, and beam diameter as inputs, demonstrating that data-driven models can surpass the Rosenthal analytical solution in accuracy \cite{Raihan2025, Meng2020, Akbari2022}. Subsequent works broadened input sets to include layer thickness, powder characteristics, and material properties -- density, specific heat, thermal conductivity, and melting temperature -- with multi-material studies across 316L stainless steel, Ti-6Al-4V, and AlSi10Mg showing accuracy gains when physics-inspired material descriptors were incorporated alongside process parameters \cite{Mosallanejad2024, Rahman2024}. The MeltpoolNet framework showed that adding absorptivity coefficients to process parameters improved cross-dataset prediction of melt pool width and depth \cite{Akbari2022}, while physics-informed deep learning approaches have been applied to molten pool morphology prediction in directed energy deposition \cite{Cao2024}. For melting regime classification, ML models trained on process parameters have been used to distinguish conduction, transition, and keyhole conditions, enabling data-driven process map construction \cite{Scime2019}. Faegh et al. (2025)  identified physics-based feature engineering as one of three principal strategies for incorporating domain knowledge into ML models -- alongside physics-constrained architectures and physics-based loss functions -- and highlighted cross-material generalization as a central open challenge \cite{Faegh2025}. Despite this progress, a consistent limitation remains: models trained on a single material generalize poorly to other alloys, and studies conducted on a systematic, controlled benchmark comparing physics-informed dimensionless input representations against raw process parameters across multiple ML algorithms and material systems with rigorous grouped evaluation remain largely underexplored \cite{Vendrell-Gallart2025, Kim2023}.

To address the generalization challenge, the authors utilized the dimensionless numbers derived from first principles of laser-material thermal interaction.
The normalized enthalpy, introduced by Hann et al. (2011) \cite{Hann2011} and evaluated by King et al. (2015) \cite{King2015} specifically for LPBF conditions, encodes the ratio of deposited laser energy density to the enthalpy of melting -- a material-agnostic quantity that absorbs the thermophysical properties of the alloy directly into the feature. The Peclet number represents the ratio of convective to conductive heat transport, characterizing the relative influence of scan speed and thermal diffusivity on melt pool shape \cite{Doring2025}. Critically, both numbers are computable from independently measurable quantities -- laser parameters and material properties -- without any melt pool observations, preserving their role as genuine predictive inputs rather than response-derived descriptors. Analytical studies have demonstrated that plotting normalized melt pool depth and width against normalized enthalpy and Peclet number collapses data from diverse alloys onto common trend lines -- an outcome that requires individual calibration curves for each material when raw process parameters are used instead \cite{Naderi2023, Hann2011, Parsazadeh2024}. A dimensionless scaling law for melt pool width in LPBF further demonstrated that dimensional analysis-based representations can characterize melt pool width in a beam-size-independent, material-spanning manner \cite{Yang2022}. Physics-informed ML for LPBF wall thickness prediction has also shown that models incorporating dimensionless numbers achieve lower prediction error than purely data-driven counterparts while requiring fewer training points \cite{PenaCorredor2025}. Nevertheless, no prior work has used normalized
enthalpy $\Delta H/h_\mathrm{s}$ and the Peclet number $Pe$ as direct ML feature inputs in a systematic, multi-model, multi-material regression and classification benchmark to the best of the authors' knowledge -- a gap that limits the field's understanding of when and why physics-informed representations outperform raw process parameters in practical data-limited scenarios.
 
Alongside input representation, the data scarcity inherent to LPBF experiments motivates evaluation of ML architectures that are specifically designed to perform well with limited training data \cite{enayet2026data}. Gaussian process regression has been the conventional choice in this setting because of its probabilistic predictions and sample efficiency, and it has been applied to melt pool geometry and process property prediction across several LPBF systems \cite{Naderi2023, Kim2023}. However, GPR performance depends critically on kernel selection and hyperparameter optimization, and isotropic kernels may inadequately capture the anisotropic sensitivity of melt pool dimensions to inputs spanning very different physical scales -- a limitation that becomes pronounced when training datasets are small and input ranges are wide \cite{Sudeep2026}. A fundamentally different approach has recently emerged in the form of pretrained Tabular Foundation Models (TFMs), which learn a general prior over tabular regression and classification tasks by training on millions of datasets, then apply that on new datasets without dataset-specific gradient updates \cite{Hollmann2025}. TabPFN demonstrated that this in-context learning paradigm outperforms all prior ML methods -- including gradient-boosted trees tuned for four hours -- on datasets with up to 10,000 samples, making accurate predictions in seconds \cite{Hollmann2025}. Related models TabDPT \cite{Ma2025} and TabICLv2 \cite{Qu2026} extend this paradigm with improved scalability and architectural refinement. Given the importance of data-efficient learning strategies for advanced manufacturing, pretrained TFMs represent a concrete recent realization of this direction. While TFMs have begun to appear in materials discovery and biomedical applications \cite{Hu2026}, their application to LPBF process-property modeling with data-scarce environments has been limited, leaving open the question of whether synthetic pretraining priors transfer effectively to the structured, physics-governed relationships of laser-based manufacturing processes.
 
This study addresses three connected gaps in the literature. First, while dimensionless numbers have been applied analytically to melt pool scaling, they have not been evaluated as direct ML input features in a controlled, multi-model benchmark. Second, the generalization advantage of physics-informed dimensionless representations relative to dimensional process parameters has not been quantified empirically across multiple material systems with rigorous, group-based data partitioning that prevents process-condition leakage. Third, pretrained tabular foundation models have not been evaluated on LPBF process-property prediction, leaving their transferability to manufacturing contexts unestablished. The specific contributions of this work are the following:
 
\begin{enumerate}
\setlength{\itemsep}{2pt}
    \item A systematic benchmark comparing dimensionless inputs against process parameters and material properties for melt pool and depth and width regression and melting mode classification across three material systems -- 316L stainless steel, IN625, and IN718 alloys -- and their combination;
 
    \item Evaluation of pretrained tabular foundation models (TabPFN, TabDPT, TabICLv2) alongside established baselines (GPR, SVR, random forest, XGBoost) for LPBF melt pool prediction;
 
    \item Empirical demonstration that dimensionless inputs substantially improve cross-material generalization, providing quantitative ML-level evidence for a hypothesis previously supported only by analytical scaling arguments; and
 
    \item Identification of the physical boundary conditions under which dimensionless representations outperform process parameters and material properties as inputs -- and the cases where they do not -- with mechanistic interpretation.
\end{enumerate}

Together, these findings offer a principled, evidence-based framework for designing generalizable, data-efficient ML systems for LPBF process-window identification and qualification, with direct implications for smart manufacturing systems operating across diverse materials and processing conditions. The remainder of this paper discusses the methodology, including the datasets and prediction models, in Section \ref{Methodology}, presents the findings in Section \ref{Result and Discussion}, and concludes in Section \ref{Conclusion}.

\begin{figure*}[!htb]
    \centering
    \includegraphics[width=.95\linewidth]{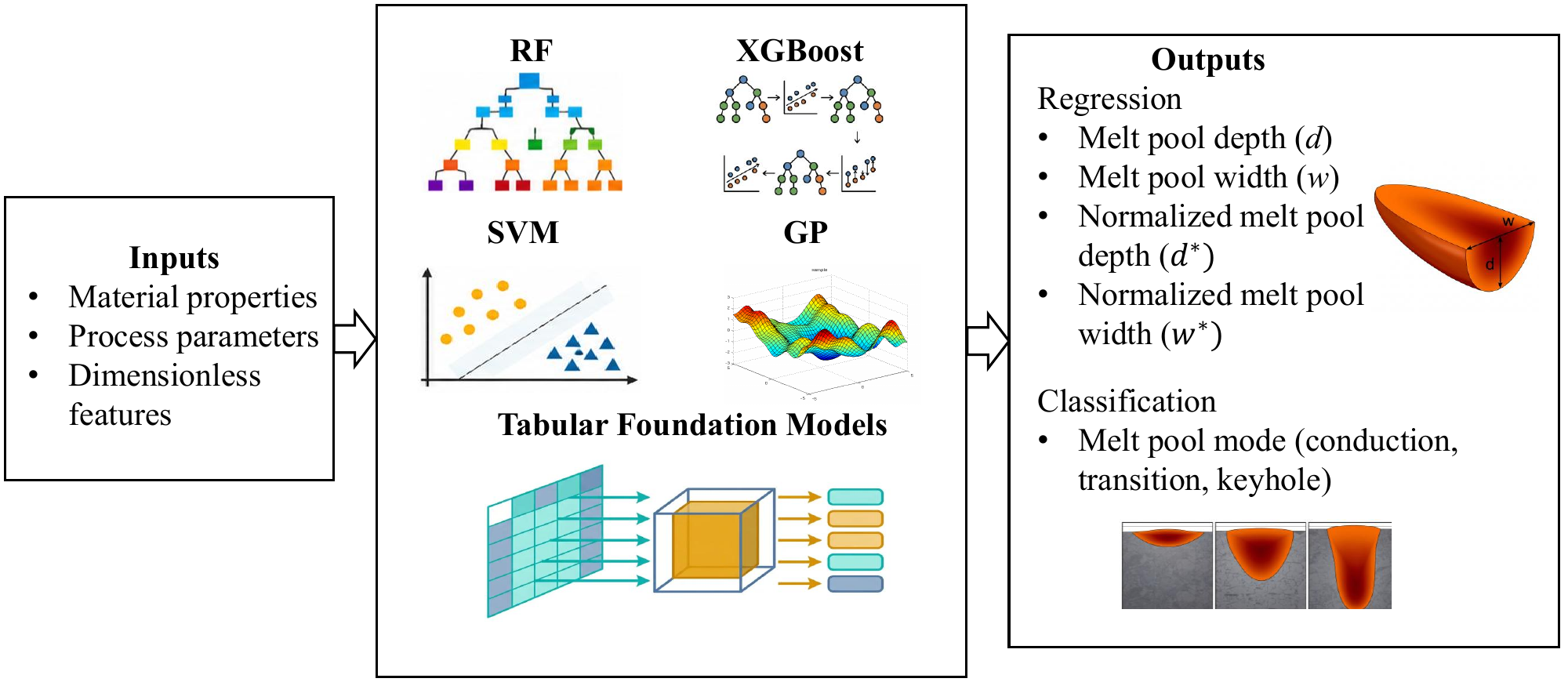}
    \caption{Input features, prediction models, and predicted outputs in the workflow.}
    \label{fig:workflow}
\end{figure*}

\section{Methodology}
\label{Methodology}

Two experimental datasets were harmonized into a common schema, and two
distinct feature representations were constructed: one using dimensional
process parameters with material properties, and one using
physics-informed dimensionless parameters. When dimensional inputs were
used, the regression targets were the directly measured melt pool depth ($d$) and width
($w$) in physical units. When dimensionless inputs were
used, both the inputs and the targets were kept dimensionless for
internal consistency, where the melt pool depth and width were
normalized by the laser spot radius to formulate normalized melt pool depth ($d^*$) and width ($w^*$) respectively. This choice ensures that the model
operates entirely within a dimensionless space, which is a fundamental
requirement of dimensional analysis: a physically consistent model
expressed in dimensionless variables must have both its inputs and
outputs in dimensionless form, so that the functional relationship is
independent of the system of units and can be generalized across process
conditions and materials \cite{Naderi2023,Hann2011}. Seven machine learning models
spanning tree ensembles, kernel methods, and pretrained tabular
foundation models were evaluated under both representations for
regression of melt pool width and depth and for classification of the
prevailing melt pool mode. Figure \ref{fig:workflow} summarizes the workflow.

\subsection{Experimental Data}
\label{Experimental Data}
Two single-track experimental datasets were used. The
first was reported by Hofmann et al. (2026) \cite{Hofmann2026}, providing 316L
stainless steel single-track measurements under systematic variation of
laser power, scan speed, spot diameter, and powder layer thickness. A
working subset of 210 observations with no layer thickness was used. The authors referred to the dataset as `316L' later in this study. The second dataset was
collected by Weaver et al. (2023) \cite{WeaverJordanS.DeisenrothDavidMekhontsevSergeyHeigelJarredLane2023} and published by NIST (National
Institute of Standards and Technology) from laser powder bed fusion (LPBF)
experiments on bare plates of IN625 and IN718 alloys, conducted across
two LPBF platforms while systematically varying laser power, scan speed,
and spot diameter. Tracks were cross-sectioned, metallographically
prepared, and chemically etched, after which melt pool width and depth
were measured from optical micrographs, where width is the maximum
lateral extent and depth is the maximum subsurface dimension of the
solidified track. All 250 observations were used after applying
complete-case filtering across all required features and targets. The second dataset has been referred to as `IN625+IN718'.
Concatenation of both working datasets yielded a combined dataset of 460
observations used for the multi-material benchmark, which has been referred to as the `Combined' dataset.
 
Units were harmonized before modeling and numeric conversion was applied
to all required predictors and targets. The paired regression
benchmark enforced complete cases across both feature representations
and both geometric targets, ensuring that all models were evaluated on
an identical set of observations. Spot diameter was verified to be
strictly positive, and normalized geometric targets were audited against
their corresponding physical measurements to confirm consistency. Repeat
measurements at the same process condition were retained as independent
observations and assigned to the same process-condition group for
splitting purposes.

\subsection{Featurization}
\label{Featurization}
Two feature representations were constructed and compared systematically
across all models. The following subsections discuss the features and labeling of the melt pool modes.

\begin{table*}[h]
\caption{Features employed in prediction models.}
\label{tab:input features}
\small 
\begin{tabular*}{\hsize}{@{\extracolsep{\fill}}lll@{}}
\toprule
Process Parameters (unit) & Material Properties (unit) & Dimensionless Parameters \\
\midrule
Laser Power, $P$ (W)                    & Density, $\rho$ (kg\,m$^{-3}$)                        & Normalized Enthalpy, $\Delta H/h_\mathrm{s}$ \\
Scan Speed, $u$ (mm/s)                  & Specific Heat Capacity, $c_p$ (J\,kg$^{-1}$\,K$^{-1}$) & Peclet Number, $Pe$                      \\
Laser Spot Diameter, $D_b = 4\sigma$ ($\mu$m) & Thermal Conductivity, $\kappa$ (W\,m$^{-1}$\,K$^{-1}$) &                                              \\
                                        & Melting Temperature, $T_m$ (K)                         &                                              \\
\hline
\end{tabular*}
\end{table*}

\subsubsection{Process Parameters and Material Properties}
\label{Process Parameters and Material Properties}

The dimensional process parameters representation used laser power, scan
speed, and estimated laser spot diameter as the primary process inputs.
Laser spot diameter was included as an explicit predictor because the
spatial concentration of the heat source is not recoverable from power
and speed alone; its influence on melt pool dimensions and regime
boundaries has been demonstrated experimentally for both conduction and
keyhole conditions \cite{Schlenger2025,weaver2022laser}. For the IN625+IN718 and the combined
datasets, four thermophysical material descriptors were added: density,
specific heat capacity, thermal conductivity, and melting temperature.
These quantities govern heat storage, conduction, and the temperature
scale for melting, and are required to distinguish the response of
different alloys from the same process settings \cite{Mosallanejad2024,Akbari2022}. For the
single-material 316L, material descriptors are constant across
observations and do not contribute information; the learning models therefore
used the three process variables only. The process parameters (PP) and material properties (MP) -- used as dimensional input -- are listed in Table \ref{tab:input features} and the specific values used for the material properties are listed in Table \ref{tab:material properties}. For the combined multi-material
benchmark, material identity was additionally encoded as a categorical
predictor using an encoder fitted exclusively to the training partition,
consistent with established practice in multi-material melt pool
modeling \cite{Akbari2022}.

\begin{table}[h]
\caption{Material properties used in the study \cite{Hofmann2026,Naderi2023}.}
\label{tab:material properties}
\small
\begin{tabular*}{\hsize}{@{\extracolsep{\fill}}lccc@{}}
\toprule
Property (unit) & 316L & IN625 & IN718 \\
\midrule

$\rho$ (kg\,m$^{-3}$)                & 7980 & 8440    & 7727 \\
$c_p$ (J\,kg$^{-1}$\,K$^{-1}$)      & 855  & 720     & 650  \\
$\kappa$ (W\,m$^{-1}$\,K$^{-1}$)    & 36.7 & 28      & 29   \\
$T_m$ (K)                            & 1697 & 1563.15 & 1526 \\
\hline
\end{tabular*}
\end{table}

\subsubsection{Dimensionless Parameters}
\label{Dimensionless Parameters}

Dimensionless parameters are unitless combinations of numbers of dimensional process
parameters and thermophysical material properties, formed such that all
units cancel, enabling the description of physical phenomena in a
material- and machine-independent form. Derived through the
Vaschy--Buckingham $\pi$-theorem, these parameters allow experimental data
from different materials, laser systems, and processing conditions to be
collapsed onto unified relationships, thereby facilitating the transfer
of process knowledge across materials and machines \cite{weaver2022laser}. In LPBF, the
dimensionless representation encodes process conditions and material
response through two physics-informed quantities: the normalized
enthalpy, $\Delta H/h_\mathrm{s}$ and the Peclet number, $Pe$ as listed in Table \ref{tab:input features}. With a
Gaussian beam convention $D_b = 4\sigma$ and beam radius $r = 2\sigma$,
the thermal diffusivity $\alpha$ and the melting enthalpy
$h_\mathrm{s}$ are defined as:
\begin{equation}
    \alpha = \frac{\kappa}{\rho\, c_p} 
    \label{eq:alpha}
 \end{equation}   

\begin{equation}
    h_\mathrm{s} = \rho\, c_p\,(T_m - T_0)
    \label{eq:hs}
\end{equation}
The two dimensionless features are then:
\begin{equation}
    \frac{\Delta H}{h_\mathrm{s}} = \frac{A\,P}{h_\mathrm{s}\,\sqrt{\pi\,\alpha\,u\,\sigma^3}}
    \label{eq:normalized_enthalpy}
\end{equation}

\begin{equation}
    Pe = \frac{u\,(2\sigma)}{\alpha}
    \label{eq:Peclet}
\end{equation}
Here $A$ is the laser absorptivity, set to $A = 0.4$ for all materials,
consistent with the value adopted in prior LPBF scaling studies
\cite{Naderi2023,King2015}; $T_0$ is the initial substrate temperature; $\sigma$
is the Gaussian spatial standard deviation of laser spot size; and $\Delta H$ is the absorbed energy density. The
computed dimensionless features were verified against stored values in
the working files for each dataset.
 
$\Delta H/h_\mathrm{s}$ in Eq.(\ref{eq:normalized_enthalpy}) measures the supplied absorbed power relative to
the energy required to melt the material. Hann et al. (2011) \cite{Hann2011} and King et al. (2015) \cite{King2015} showed that normalized melt pool depths  ($d^*$) for different alloys collapse onto a single trend line when plotted against $\Delta H/h_\mathrm{s}$. $Pe$ in Eq. (\ref{eq:Peclet}) captures the ratio of heat-source translation to
thermal diffusion over the beam-radius scale and governs the melt pool
shape transition from conduction to keyhole \cite{Doring2025}. At low $Pe$, the melt pool is symmetric and roughly circular as heat has time to diffuse in all directions. At high $Pe$, the melt pool elongates in the scan direction because the heat source moves faster than conduction can redistribute the energy. Together, the pair
provides a physically motivated representation that combines material
response, source geometry, and operating conditions into two scalar
features, independent of the alloy system \cite{Yang2022,PenaCorredor2025}. A constant
absorptivity remains an approximation because energy coupling varies
with surface state, melt regime, and keyhole depth; this is acknowledged
as a limitation.

\subsubsection{Melt Pool Mode Labels}
\label{Melt Pool Regime Labels}

Regime classification used geometric labels derived from the melt pool
aspect ratio $d/w$. Following the convention supported by King et al.\
\cite{King2014} and Hofmann et al.\ \cite{Hofmann2026}, tracks with $d/w \leq 0.5$ were
labeled as conduction mode, those with $0.5 < d/w \leq 0.8$ as transition
mode, and those with $d/w > 0.8$ as keyhole mode. Measured depth, width,
their normalized forms ($d^*$, $w^*$), and the aspect ratio were
explicitly excluded from classifier input features, so that regime
categories were predicted solely from process and material information.

\subsection{Dataset Splitting and Metrics}
\label{Dataset Splitting and Metrics}

 The process-condition group identifier was constructed from the tuple
(Material, $P$, $u$, $D_b$) for each observation. All observations
sharing this tuple were assigned to the same partition in every split,
ensuring that repeated measurements of the same process condition cannot
appear on both sides of a train-evaluation boundary. This grouped
splitting strategy prevents the optimistic bias that arises when
near-identical conditions are split across folds, and mirrors the
intended deployment scenario where the model must predict an unseen
process setting rather than interpolate among replicates of a known one.
 
A single grouped holdout partition was created using
\texttt{GroupShuffleSplit}, reserving 20\% of the unique
process-condition groups for holdout evaluation. Because group sizes
vary, the corresponding fraction of individual observations may differ
from 20\%. Disjointness between the training and holdout groups was
verified explicitly by asserting that the sets of group identifiers in
each partition do not intersect. An independent five-fold grouped
cross-validation (CV) was conducted using \texttt{GroupKFold} over all
cleaned observations, including those assigned to the holdout in the
separate split. The holdout and CV evaluations are therefore fully
independent analyses: the holdout was not used for model selection or
hyperparameter tuning, and CV results describe generalization across the
full dataset rather than performance on the withheld groups alone. Group
disjointness was verified at each CV fold using the same assertion.
The hyperparameters were fixed prior to all evaluations.
 
Regression performance was quantified using the coefficient of
determination ($R^2$), mean absolute percentage error (MAPE, \%), and
range-normalized RMSE (NRMSE). Higher $R^2$ values indicate better performance, and lower MAPE and NRMSE values indicate better performance. The melt pool depth and width
metrics were computed separately. For the dimensionless representation,
native dimensionless metrics describe the normalized regression task,
while reconstructed physical-unit metrics allow direct comparison with
raw-input predictions. Classification performance was quantified using
accuracy, recall, macro F1, and precision matrices.

\subsection{Data Leakage}

Group disjointness between fitting and evaluation partitions was
verified explicitly at each split. All preprocessing steps that could
carry target-side information -- including input feature scaling for
kernel methods, target standardization, and categorical encoding -- were
fitted exclusively within each training partition and applied to the
evaluation partition. The normalization audit confirmed that the
dimensionless targets $d^*$ and $w^*$ reproduced the ratio of physical
measurements to beam radius, and that no geometric target,
target-derived quantity, or aspect ratio entered the predictor matrix.
For tabular foundation model inference, labeled context observations
were drawn from the training partition only; query observations used
only their predictor values during scoring.

\subsection{Prediction Models}

The classical ML models and tabular foundation models (TFMs) evaluated on our datasets discussed previously are introduced briefly as follows. All the models have been used for both regression and classification tasks. Table \ref{tab:regressor settings} and Table \ref{tab:classifier settings} present the explicit settings used for these models.

\begin{table}[h]
\caption{Regressor settings in the current benchmark.}
\label{tab:regressor settings}
\small
\begin{tabular*}{\columnwidth}{@{\extracolsep{\fill}}lp{0.65\columnwidth}@{}}
\toprule
Model & Explicit settings \\
\midrule
Random forest    & 500 trees; remaining tree controls at defaults. \\
XGBoost          & 500 boosting rounds; depth 6; learning rate 0.05; squared-error objective; histogram tree method.           \\
GPR              & Constant $\times$ RBF + white-noise kernel; input scaling.         \\
SVR              & RBF kernel; $C=10$; $\varepsilon=0.1$; input and target standardization; $\gamma$= scale.         \\
TabPFN           & Pretrained regressor; 8 ensemble members.   \\
TabDPT           & Pretrained regressor; 8 ensembles; context size 2048; batch size 256. \\
TabICLv2         & Pretrained regressor; 8 ensemble members.            \\
\hline
\end{tabular*}
\end{table}

\begin{table}[h]
\caption{Classifier settings in the current benchmark.}
\label{tab:classifier settings}
\small
\begin{tabular*}{\columnwidth}{@{\extracolsep{\fill}}lp{0.65\columnwidth}@{}}
\toprule
Model & Explicit settings \\
\midrule
Random forest    & 500 trees; max depth unrestricted; min samples to split $=2$; min samples per leaf $=1$.                 \\
XGBoost          & 500 rounds; depth $=3$; learning rate $=0.1$; multi:softprob; tree method=hist.      \\
GPC              & StandardScaler; RBF kernel;  max\_iter\_predict=100; one-vs-rest. \\
SVC              & RBF kernel; $C=1.0$; $\gamma$ = scale;             \\
                 & probability enabled; no class weighting.                  \\
TabPFN           & Pretrained classifier; inference batch size: package defaults.                \\
TabDPT           & Pretrained classifier; inference batch size: 256.
                               \\
TabICLv2         & Pretrained classifier; inference batch size: 4.              \\
\hline
\end{tabular*}
\end{table}

\subsubsection{Random Forest}
Random Forest (RF) is an ensemble learning method that constructs a large set of decision trees, each trained on a randomly sampled subset of the training data and a randomly selected subset of input features at each split node. Final predictions are obtained by aggregating the outputs across all trees – majority voting for classification and averaging for regression – which reduces variance and overfitting compared to individual decision trees \cite{Breiman2001}

\subsubsection{XGBoost}
XGBoost (XGB) is a scalable gradient tree boosting system that builds an ensemble of decision trees sequentially, where each new tree is trained to correct the residual errors of the existing ensemble. Its efficiency stems from several algorithmic innovations, including a sparsity-aware tree learning algorithm, a weighted quantile sketch for approximate tree learning, and cache-optimized computation, allowing it to scale to large datasets significantly faster than conventional boosting methods \cite{Chen2016}

\subsubsection{Gaussian Process Model (GPR and GPC)}
Gaussian Process (GP) is a probabilistic, non-parametric Bayesian method that places a prior distribution over functions and updates it given observed training data to produce a full posterior predictive distribution, with predictions accompanied by uncertainty estimates  \cite{williams2006gaussian}. In the regression setting, GPR infers a continuous output by conditioning the GP prior on training observations using a kernel function that encodes assumptions about the smoothness and structure of the underlying function. For classification tasks, GPC applies the same GP prior with a likelihood function that maps the latent function outputs to class probabilities.

\subsubsection{Support Vector Machine (SVR and SVC)}
Support Vector Machine (SVM), introduced by Cortes and Vapnik (1995), operates by mapping input data into a high-dimensional feature space through a nonlinear kernel transformation, within which a linear decision boundary or regression function is constructed with maximum margin. For regression tasks, the model fits a function within a specified tolerance band around the training data, penalizing only predictions that fall outside this band, which makes it robust to outliers. SVMs perform well in high-dimensional spaces with relatively small training datasets – a common scenario in LPBF experimental studies \cite{Cortes1995}.

\subsubsection{TabPFN}
Tabular Prior-Data Fitted Network (TabPFN) introduced the foundational architecture for tabular foundation models. It is pre-trained once on synthetically generated datasets and, at inference time, accepts training and test samples as input and produces predictions in a single forward pass through a transformer network, effectively approximating Bayesian inference without any additional model fitting \cite{hollmann2022tabpfn}. While originally demonstrated on small classification datasets, its ability to generalize without retraining makes it conceptually attractive for LPBF process-property modeling where experimental data is scarce.

\subsubsection{TabDPT}
Tabular Discriminative Pre-trained Transformer (TabDPT) extends TabPFN by incorporating real tabular datasets into the pretraining procedure alongside self-supervised learning, showing that real-world data contains richer signal than synthetic generators alone \cite{Ma2025}. The resulting model supports both classification and regression tasks within a single shared architecture and exhibits predictable performance scaling with model and data size, suggesting its suitability for predicting continuous outputs such as melt pool dimensions across varying process conditions.

\subsubsection{TablCLv2}
Tabular In-Context Learning, version 2 (TabICLv2) further advanced the state of the art through three key innovations: a more diverse synthetic data generation engine, a scalable attention mechanism enabling generalization to larger datasets, and an improved optimization protocol \cite{Qu2026}. Together, these allow TabICLv2 to handle datasets ranging from hundreds to millions of samples efficiently, making it applicable across the full spectrum of LPBF experimental dataset sizes.

\section{Result and Discussion}
\label{Result and Discussion}

This section presents the comparative performance of seven machine
learning models evaluated under two feature representations --
physics-informed dimensionless parameters (normalized enthalpy and
P\'{e}clet number) and dimensional process parameters with material
properties -- across three datasets (316L, IN625+IN718, and
Combined). Results are reported from five-fold grouped cross-validation.

\subsection{ Melt Pool Depth Prediction}

\begin{figure*}[!htbp] 
    \centering
    
    \begin{minipage}[b]{0.33\textwidth}
        \centering
        \includegraphics[width=\linewidth]{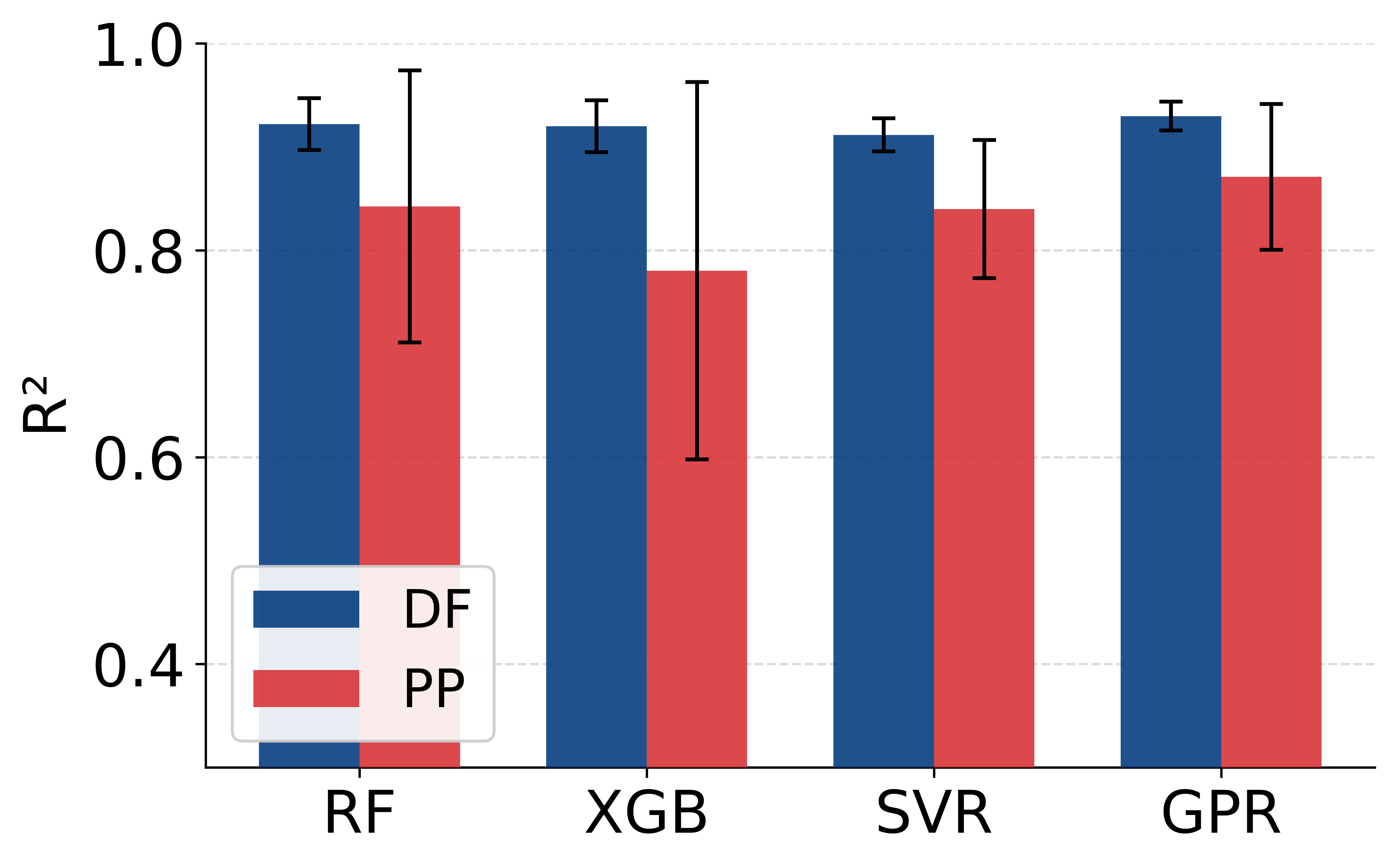}
        \subcaption{}
        \label{fig:3a}
    \end{minipage}
    \hfill
    \begin{minipage}[b]{0.33\textwidth}
        \centering
        \includegraphics[width=\linewidth]{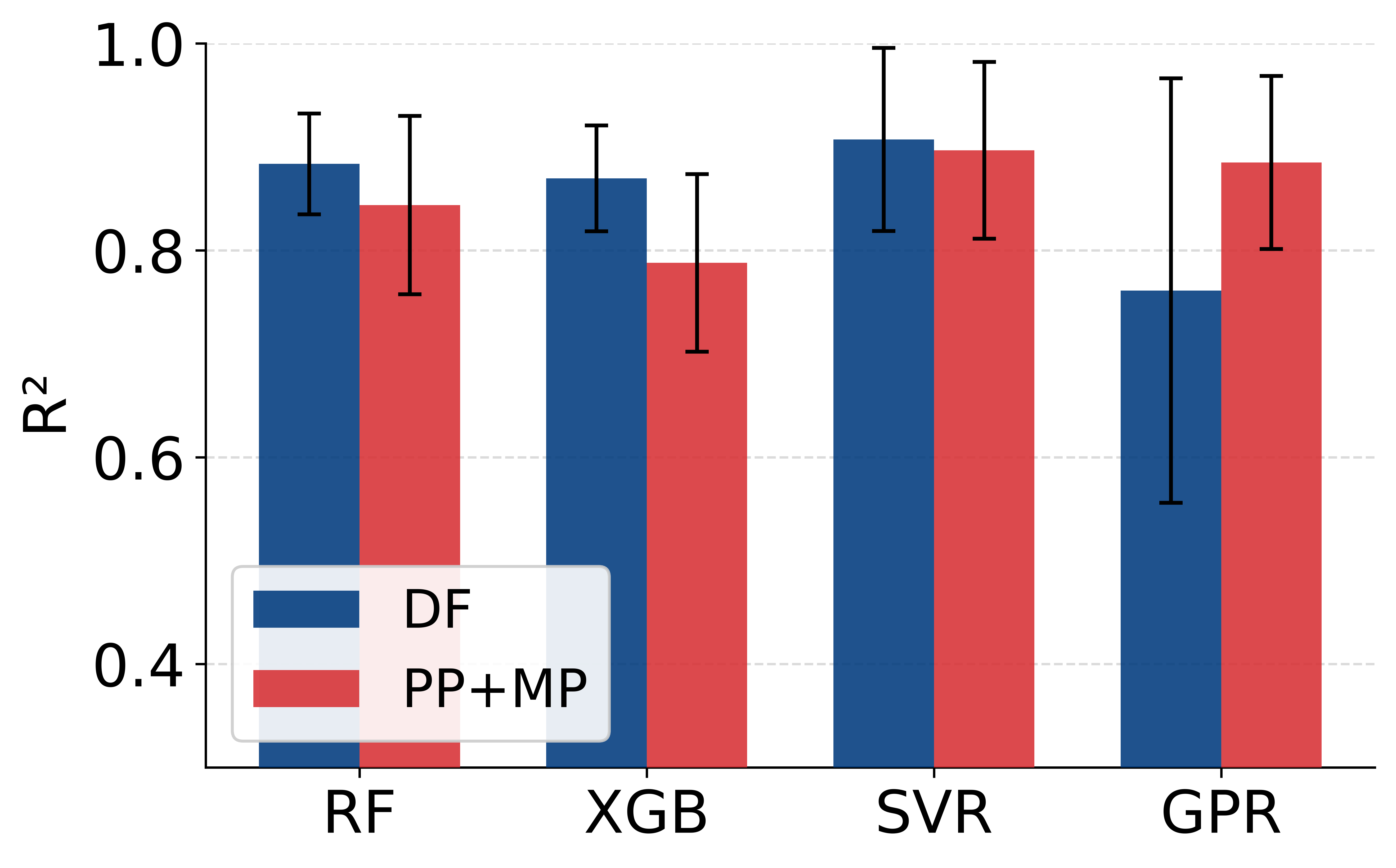}
        \subcaption{}
        \label{fig:3b}
    \end{minipage}
    \hfill      
    \begin{minipage}[b]{0.33\linewidth}
        \centering
        \includegraphics[width=\linewidth]{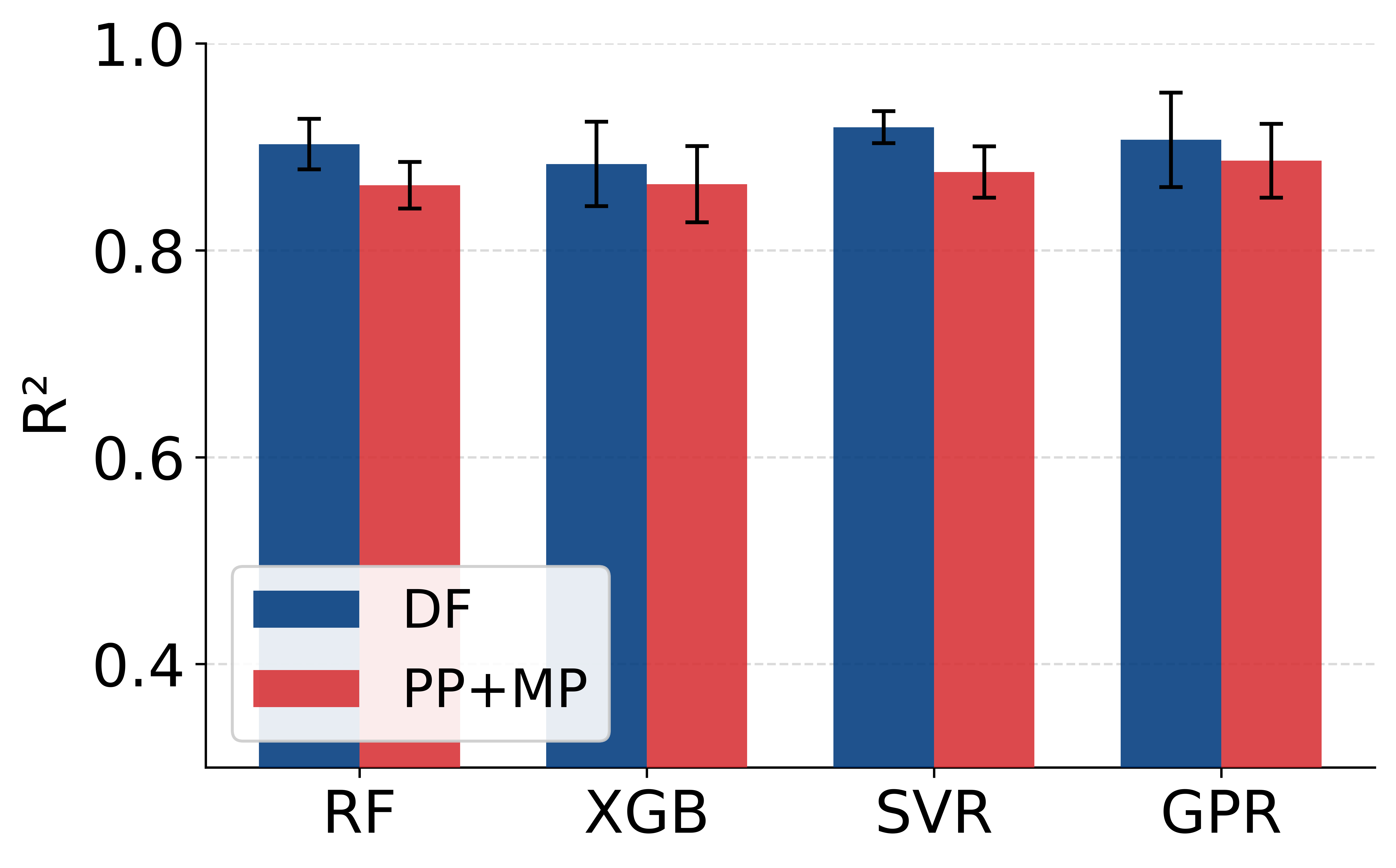}
        \subcaption{}
         \label{fig:3c}
    \end{minipage}
    \\[1ex]  

\begin{minipage}[b]{0.33\textwidth}
    \centering
    \includegraphics[width=\linewidth]{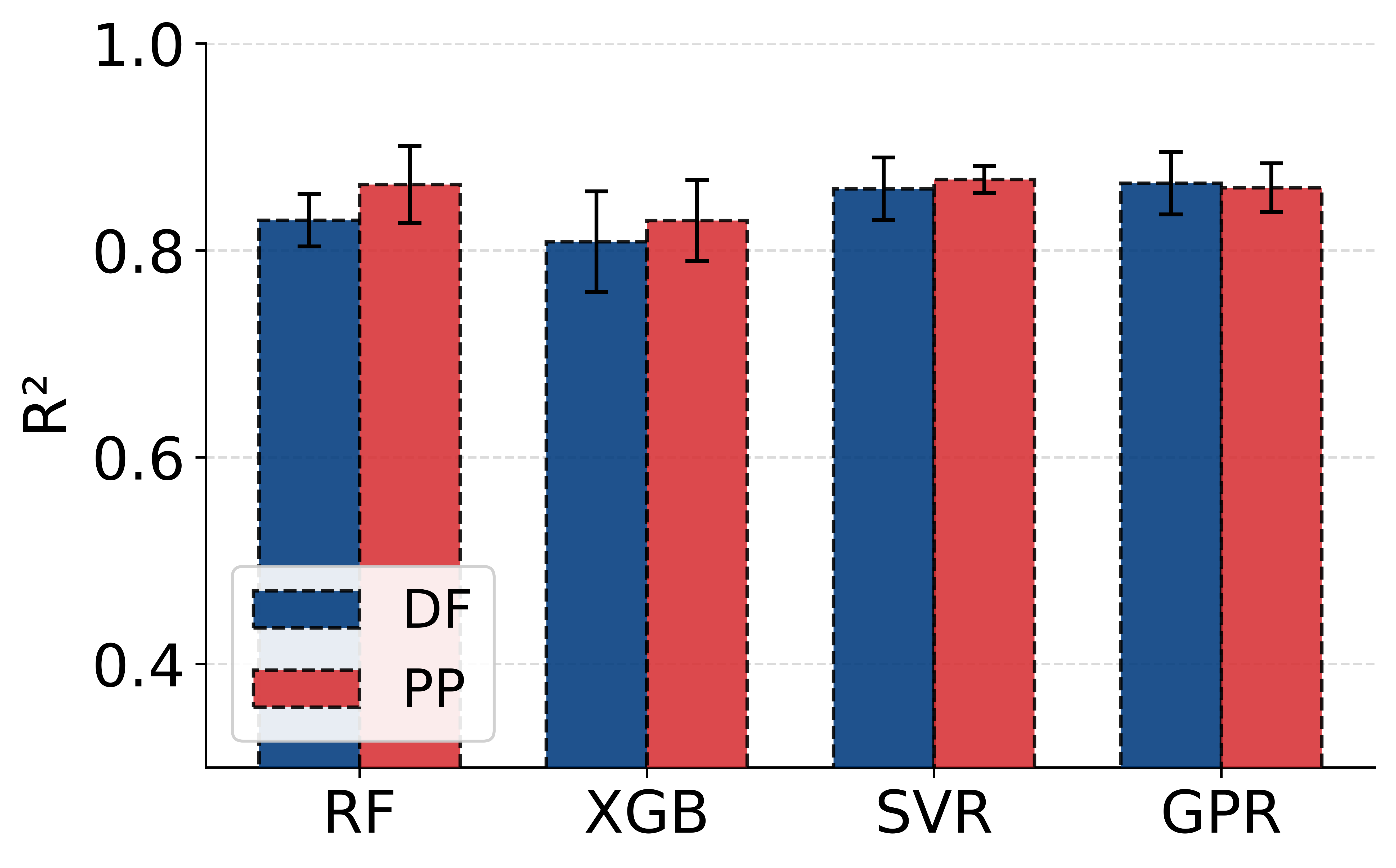}
    \subcaption{}
    \label{fig:3d}
\end{minipage}
\hfill
\begin{minipage}[b]{0.33\textwidth}
    \centering
    \includegraphics[width=\linewidth]{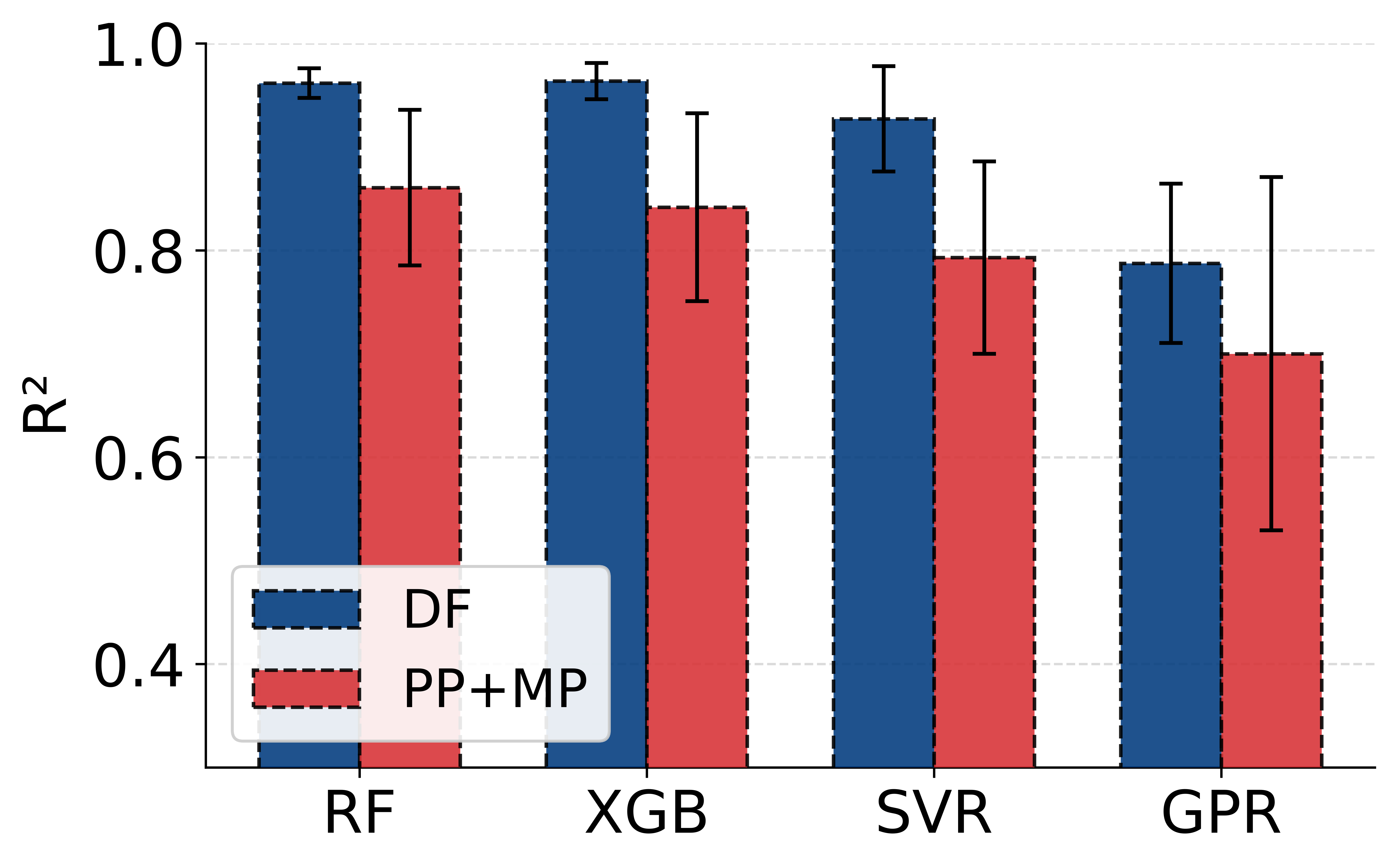}
    \subcaption{}
    \label{fig:3e}
\end{minipage}
\hfill
\begin{minipage}[b]{0.33\textwidth}
    \centering
    \includegraphics[width=\linewidth]{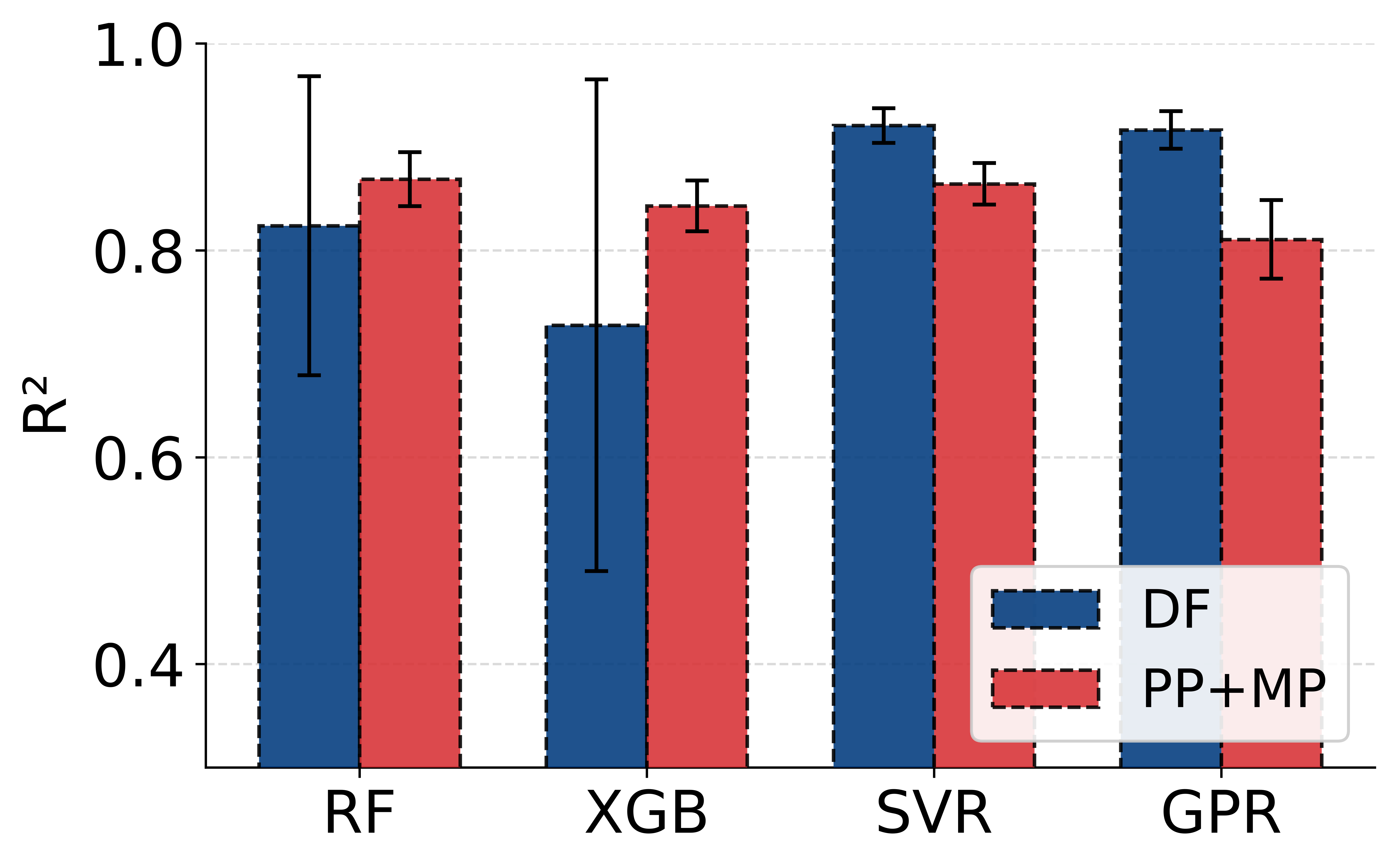}
    \subcaption{}
    \label{fig:3f}
\end{minipage}
    
    \caption{Prediction performance of melt pool depth (a,b,c) and width (d,e,f) by classical ML models, evaluated by $R^2$ values using dimensionless features (DF) and process parameters (PP) \& material properties (MP). (a) \& (d) represent performance on 316L dataset; (b) \& (e) represent performance on IN625+IN718 dataset; and (c) \& (f) represent the performance on combined dataset.}
    \label{fig:R2_classical_ml}
\end{figure*}

Figure \ref{fig:R2_classical_ml}(a,b,c),  Table \ref{tab:depth_316L}, \ref{tab:depth_nist}, and \ref{tab:depth_combined} present the performance of the melt pool depth regression results
for the 316L, IN625+IN718, and combined datasets. For melt pool depth prediction, the dimensionless representation provides a clear and consistent advantage across all three datasets. On the 316L single-material dataset, all seven models achieve higher $R^2$ under dimensionless inputs – GPR improves from 0.871 to 0.930 and XGB from 0.780 to 0.920 – alongside substantial MAPE reductions, suggesting that depth is more sensitive to the energy-material interaction encoded in normalized enthalpy than width is, as thermal penetration is the quantity most directly governed by normalized enthalpy. On the IN625+IN718 dataset, an important caveat emerges for classical kernel models: GPR and SVR exhibit anomalously high MAPE under dimensionless inputs (34.67\% and 35.14\%, respectively) despite reasonable $R^2$ values (0.761 and 0.907), indicating large proportional errors on a small subset of low-depth observations rather than systematic failure across the full range. Tabular foundation models do not exhibit this pathology – TabPFN, TabDPT, and TabICLv2 maintain MAPE of 10–12\% under the same dimensionless inputs – highlighting a practical reliability advantage of TFMs for depth prediction that $R^2$ alone does not capture. These errors should also be read against the measurement uncertainty of the IN625+IN718 melt pool depth data, whose expanded uncertainty ($k=2$) averages 14.5\% and includes an assumed 5\% along-track variability, indicating that part of the residual error reflects measurement uncertainty rather than model deficiency \cite{WeaverJordanS.DeisenrothDavidMekhontsevSergeyHeigelJarredLane2023}. The advantage of combining dimensionless inputs with tabular foundation models is most pronounced on the Combined dataset, where TabPFN achieves $R^2$ of 0.961 and NRMSE of 0.043 – the strongest result across all model-representation combinations in this study, compared to $R^2$ of 0.921 and NRMSE of 0.060 under dimensional inputs, confirming that physics-informed feature encoding and pretrained model priors together provide the most effective generalization for multi-material depth prediction as shown in Figure \ref{fig:fm_vs_ml_regression}a.

\begin{table}

\caption{Performance of classical ML models and TFMs for melt pool depth prediction on the 316L dataset.}
\label{tab:depth_316L}
\small
\begin{tabular*}{\hsize}{@{\extracolsep{\fill}}lcccc@{}}
\toprule
& \multicolumn{2}{c}{MAPE (\%)}
& \multicolumn{2}{c}{NRMSE} \\\cmidrule(lr){2-3}\cmidrule(lr){4-5}
Model & DF & PP & DF & PP \\
\midrule
RF       & 16.48 & 20.02 & 0.063 & 0.081 \\
XGB      & 18.13 & 20.22 & 0.064 & 0.096 \\
GPR      & 16.55 & 21.32 & 0.060 & 0.076 \\
SVR      & 20.53 & 23.93 & 0.068 & 0.086 \\
TabPFN   & 15.55 & 15.40 & 0.061 & 0.066 \\
TabDPT   & 15.45 & 15.01 & 0.060 & 0.062 \\
TabICLv2 & 15.84 & 14.25 & 0.056 & 0.063 \\
\hline

\end{tabular*}

\smallskip
{\small DF = Dimensionless Features; PP = Process Parameters.}
\end{table}

\begin{table}[t]
\caption{Performance of classical ML models and TFMs for melt pool depth prediction on the IN625+IN718 dataset.}
\label{tab:depth_nist}
\small
\begin{tabular*}{\hsize}{@{\extracolsep{\fill}}lcccc@{}}
\toprule
& \multicolumn{2}{c}{MAPE (\%)}
& \multicolumn{2}{c}{NRMSE} \\
\cmidrule(lr){2-3}\cmidrule(lr){4-5}
Model & DF & PP+MP & DF & PP+MP \\
\midrule
RF       & 15.97 & 13.76 & 0.100 & 0.102 \\
XGB      & 18.29 & 14.94 & 0.109 & 0.123 \\
GPR      & 34.67 & 15.93 & 0.144 & 0.084 \\
SVR      & 35.14 & 13.30 & 0.085 & 0.083 \\
TabPFN   & 10.02 &  9.53 & 0.061 & 0.081 \\
TabDPT   & 11.01 &  9.29 & 0.061 & 0.068 \\
TabICLv2 & 11.78 & 13.73 & 0.070 & 0.091 \\
\hline
\end{tabular*}

\smallskip
{\small DF = Dimensionless Features; PP = Process Parameters; MP=Material Properties.}
\end{table}

\begin{table}[h]
\caption{Performance of classical ML models and TFMs for melt pool depth prediction on the combined dataset.}
\label{tab:depth_combined}
\small
\begin{tabular*}{\hsize}{@{\extracolsep{\fill}}lcccc@{}}
\toprule
& \multicolumn{2}{c}{MAPE (\%)}
& \multicolumn{2}{c}{NRMSE} \\\cmidrule(lr){2-3}\cmidrule(lr){4-5}
Model & DL & PP+MP & DL & PP+MP \\
\midrule
RF      & 15.55 & 17.67 & 0.071 & 0.078 \\
XGB     & 16.13 & 16.04 & 0.079 & 0.077 \\
GPR     & 16.42 & 17.35 & 0.070 & 0.069 \\
SVR     & 28.76 & 17.11 & 0.064 & 0.075 \\
TabPFN  & 11.28 & 11.89 & 0.043 & 0.060 \\
TabDPT  & 12.11 & 11.55 & 0.047 & 0.057 \\
TabICLv2 & 12.09 & 13.50 & 0.045 & 0.064 \\
\hline
\end{tabular*}

\smallskip
{\small DF = Dimensionless Features; PP = Process Parameters; MP=Material Properties.}
\end{table}

\subsection{Melt Pool Width Prediction}
\label{Meltpool Width}

\begin{figure*}[!htbp] 
    \centering
    
    \begin{minipage}[b]{0.44\textwidth}
        \centering
        \includegraphics[width=\linewidth]{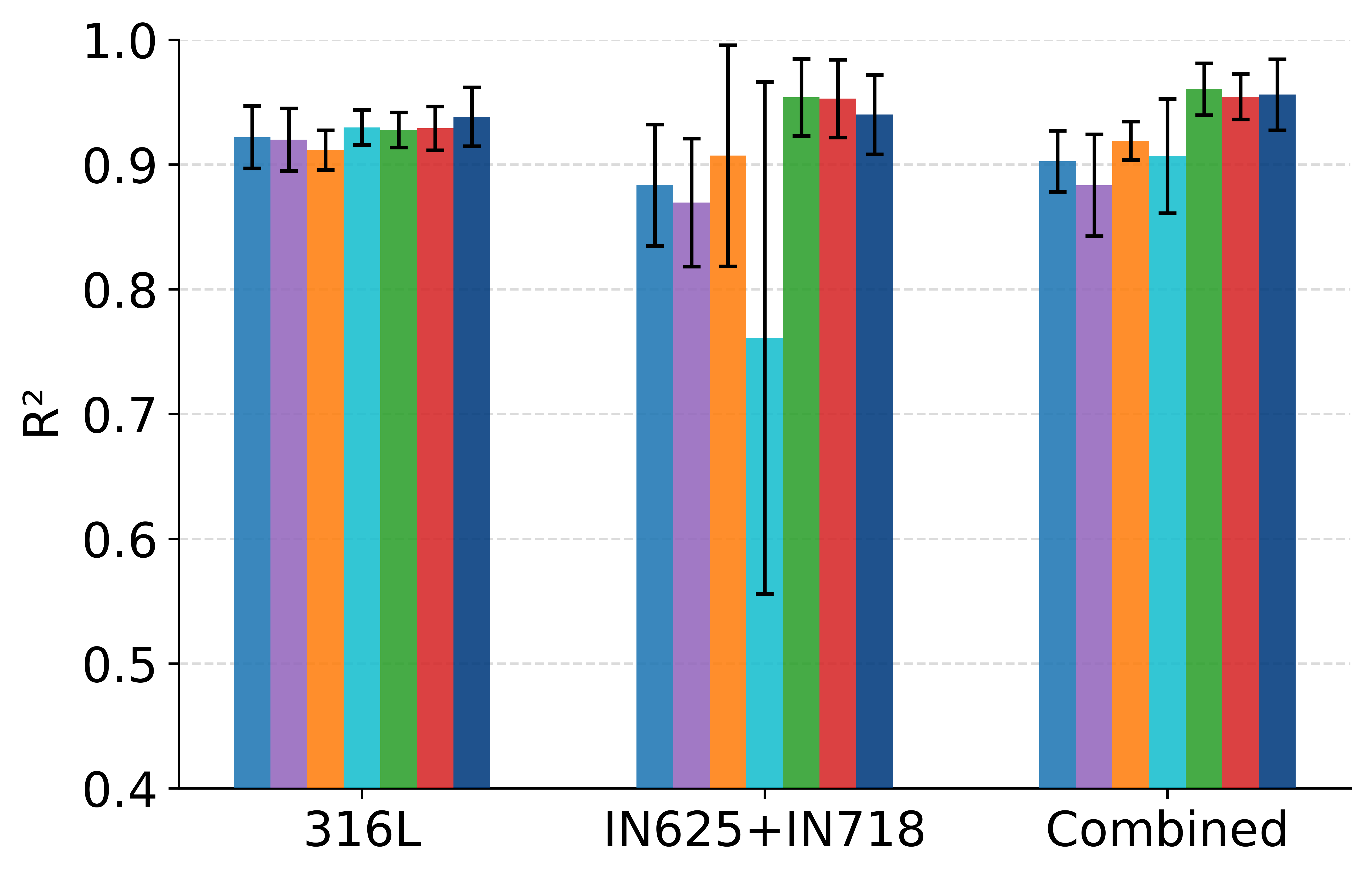}
        \subcaption{}
        \label{fig:4a}
    \end{minipage}
    \begin{minipage}[b]{0.44\textwidth}
        \centering
        \includegraphics[width=\linewidth]{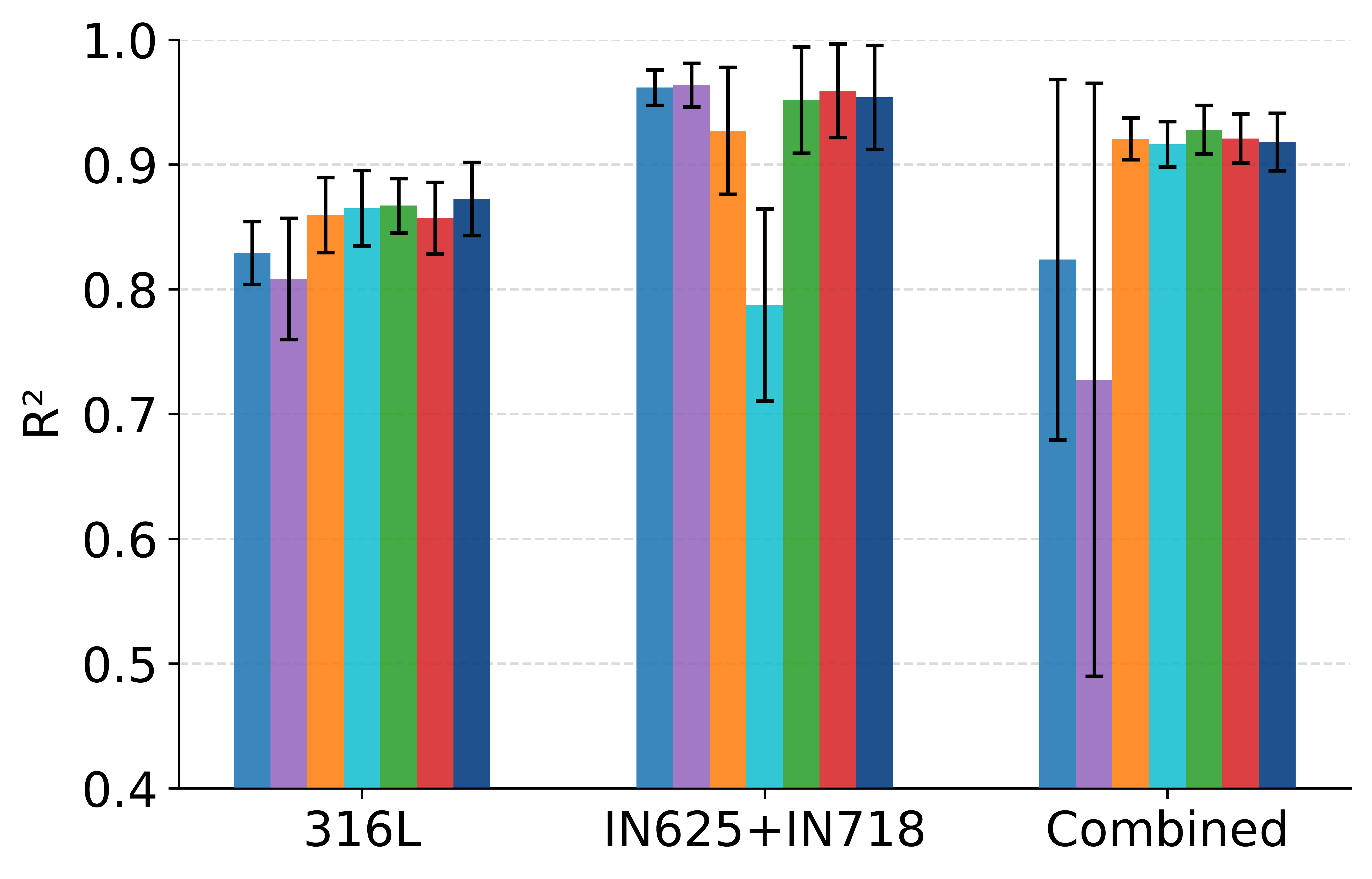}
        \subcaption{}
        \label{fig:4b}
    \end{minipage}    
    \begin{minipage}[b]{0.11\textwidth}
        \raggedleft
        \includegraphics[width=\linewidth,height=5cm, keepaspectratio=true]{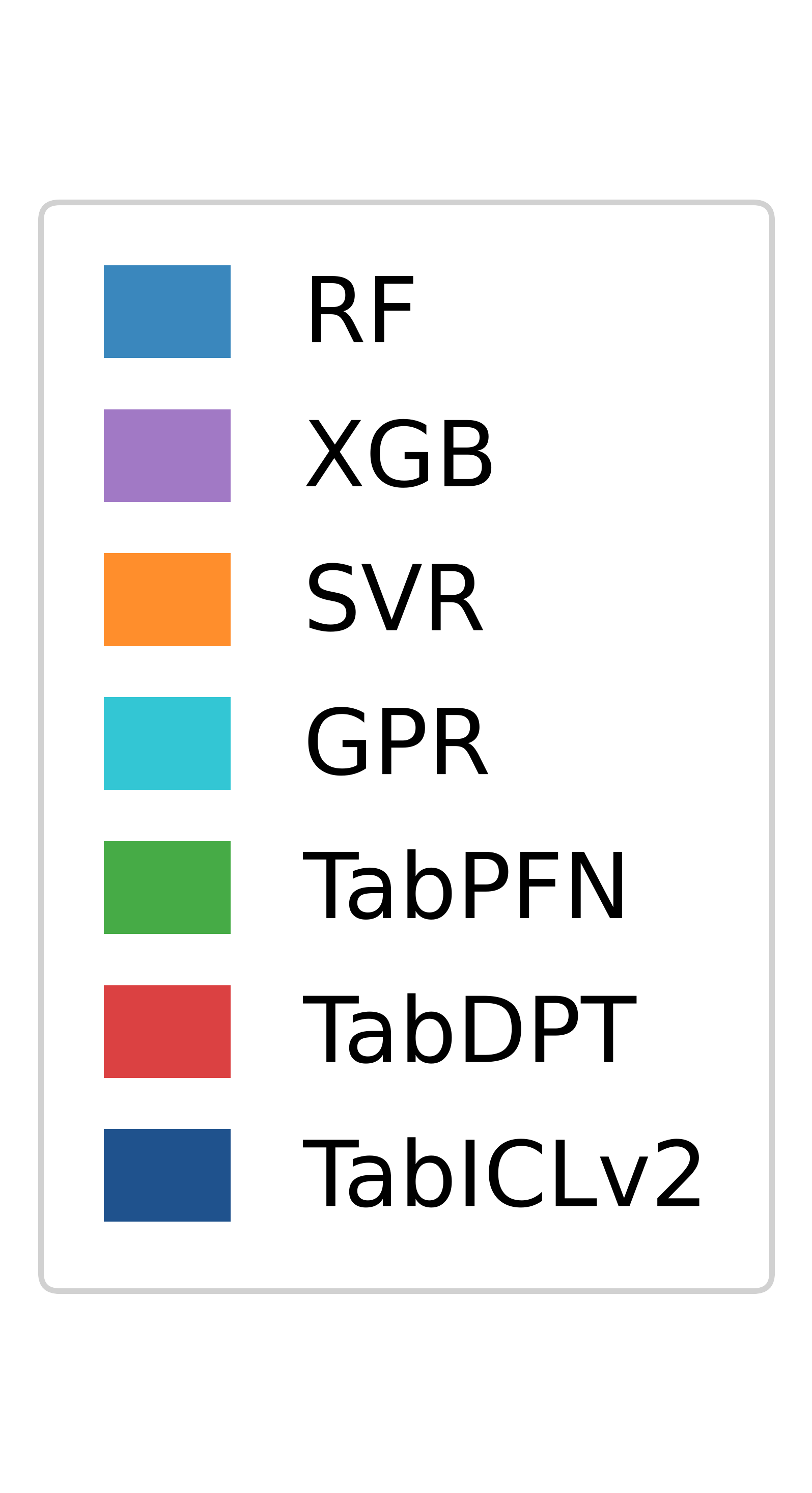}
        \label{fig:4_}
    \end{minipage}
     \caption{Predictive performance of classical ML models and TFMs for melt pool (a) depth and (b) width using dimensionless features, measured by $R^2$}
    \label{fig:fm_vs_ml_regression}

\end{figure*}

Figure \ref{fig:R2_classical_ml}(d,e,f), Table~\ref{tab:width_316L}, \ref{tab:width_nist}, and
\ref{tab:width_combined} present the performance of the melt pool width regression results
for the 316L, IN625+IN718, and Combined datasets. For melt pool width prediction, the benefit of dimensionless inputs is dataset-dependent and reveals a physically informative boundary condition. On the single-material 316L dataset, the advantage is modest and model-dependent – tree-based models (RF, XGB) marginally favor dimensional inputs, while kernel models (GPR, SVR) show negligible differences – suggesting that within a single alloy with limited process parameter variation, dimensional inputs retain sufficient discriminative information and normalization provides little added value. The picture changes substantially on the IN625/IN718 dataset, where dimensionless inputs deliver consistent and large improvements across all models: $R^2$ improves from 0.700 to 0.788 for GPR, from 0.793 to 0.927 for SVR, and from 0.861 to 0.962 for RF, with NRMSE dropping 20\%–40\% across the board – gains that reflect the material-agnostic nature of normalized enthalpy and Pe, which absorb the thermophysical differences between IN625 and IN718 by construction. On the Combined dataset, kernel-based models show substantial improvement in $R^2$ (GPR: 0.811 to 0.916; SVR: 0.864 to 0.921), while XGB degrades under dimensionless inputs (0.843 to 0.728), an outlier likely attributable to the fixed shallow tree depth being insufficient to resolve the finer splits required across the broader multi-material dimensionless input space. Tabular foundation models perform consistently well under both representations as presented in Figure \ref{fig:fm_vs_ml_regression}b, with TabPFN (0.928) and TabDPT (0.921) showing a modest but reliable advantage for dimensionless inputs, confirming that the representation benefit is most pronounced for classical models that lack the inductive bias to compensate for material heterogeneity in dimensional input space.

Within the dimensionless representation, tabular foundation models are competitive with or superior to classical methods on most datasets. On the combined dataset, TabPFN achieves the highest $R^2$ among all models (0.928), with TabDPT (0.921) and TabICLv2 (0.918) close behind. Overall, TabICLv2 and TabPFN are the most consistently strong performers for melt pool width prediction across all datasets and representations.

\begin{table}[h]
\caption{Performance of classical ML models and TFMs for melt pool width prediction on the 316L dataset.}
\label{tab:width_316L}
\small
\begin{tabular*}{\hsize}{@{\extracolsep{\fill}}lcccc@{}}
\toprule
& \multicolumn{2}{c}{MAPE (\%)}
& \multicolumn{2}{c}{NRMSE} \\
\cmidrule(lr){2-3}\cmidrule(lr){4-5}
Model & DL & PP & DL & PP \\
\midrule
RF       & 10.56 & 10.05 & 0.098 & 0.079 \\
XGB      & 11.62 & 11.08 & 0.104 & 0.089 \\
GPR      &  8.92 & 10.11 & 0.087 & 0.081 \\
SVR      &  8.71 &  9.48 & 0.089 & 0.078 \\
TabPFN   &  8.82 &  8.38 & 0.086 & 0.068 \\
TabDPT   &  9.12 &  8.50 & 0.090 & 0.069 \\
TabICLv2 &  9.21 &  8.77 & 0.085 & 0.071 \\
\hline
\end{tabular*}

\smallskip
{\small DF = Dimensionless Features; PP = Process Parameters.}
\end{table}

\begin{table}[h]
\caption{Performance of classical ML models and TFMs for melt pool width prediction on the IN625+IN718 dataset.}
\label{tab:width_nist}
\small
\begin{tabular*}{\hsize}{@{\extracolsep{\fill}}lcccc@{}}
\toprule
& \multicolumn{2}{c}{MAPE (\%)}
& \multicolumn{2}{c}{NRMSE} \\
\cmidrule(lr){2-3}\cmidrule(lr){4-5}
Model & DL & PP+MP & DL & PP+MP \\
\midrule
RF       & 4.93 & 4.64 & 0.057 & 0.076 \\
XGB      & 5.23 & 5.37 & 0.055 & 0.081 \\
GPR      & 9.56 & 5.50 & 0.138 & 0.110 \\
SVR      & 5.45 & 5.83 & 0.074 & 0.092 \\
TabPFN   & 4.45 & 4.67 & 0.057 & 0.078 \\
TabDPT   & 4.25 & 4.74 & 0.052 & 0.075 \\
TabICLv2 & 4.40 & 4.57 & 0.056 & 0.072 \\
\hline
\end{tabular*}

\smallskip
{\small DF = Dimensionless Features; PP = Process Parameters; MP=Material Properties.}
\end{table}

\begin{table}[h]
\caption{Performance of classical ML models and TFMs for melt pool width prediction on the Combined dataset.}
\label{tab:width_combined}
\small
\begin{tabular*}{\hsize}{@{\extracolsep{\fill}}lcccc@{}}
\toprule
& \multicolumn{2}{c}{MAPE (\%)}
& \multicolumn{2}{c}{NRMSE} \\
\cmidrule(lr){2-3}\cmidrule(lr){4-5}
Model & DL & PP+MP & DL & PP+MP \\
\midrule
RF       &  9.65 & 7.27 & 0.082 & 0.062 \\
XGB      & 10.76 & 7.78 & 0.101 & 0.068 \\
GPR      &  6.69 & 8.28 & 0.062 & 0.075 \\
SVR      &  6.54 & 7.50 & 0.060 & 0.064 \\
TabPFN   &  6.41 & 6.11 & 0.057 & 0.054 \\
TabDPT   &  6.57 & 6.17 & 0.060 & 0.054 \\
TabICLv2 &  6.69 & 6.32 & 0.060 & 0.055 \\
\hline
\end{tabular*}

\smallskip
{\small DF = Dimensionless Features; PP = Process Parameters; MP=Material Properties.}
\end{table}

\subsection{ Melt Pool Mode Prediction}
\label{Melt Pool Mode Prediction}

\begin{figure*}[htbp]
\centering

\begin{minipage}[t]{0.76\textwidth}
    \vspace{0pt}

    \begin{minipage}[b]{0.49\linewidth}
        \centering
        \includegraphics[width=\linewidth]{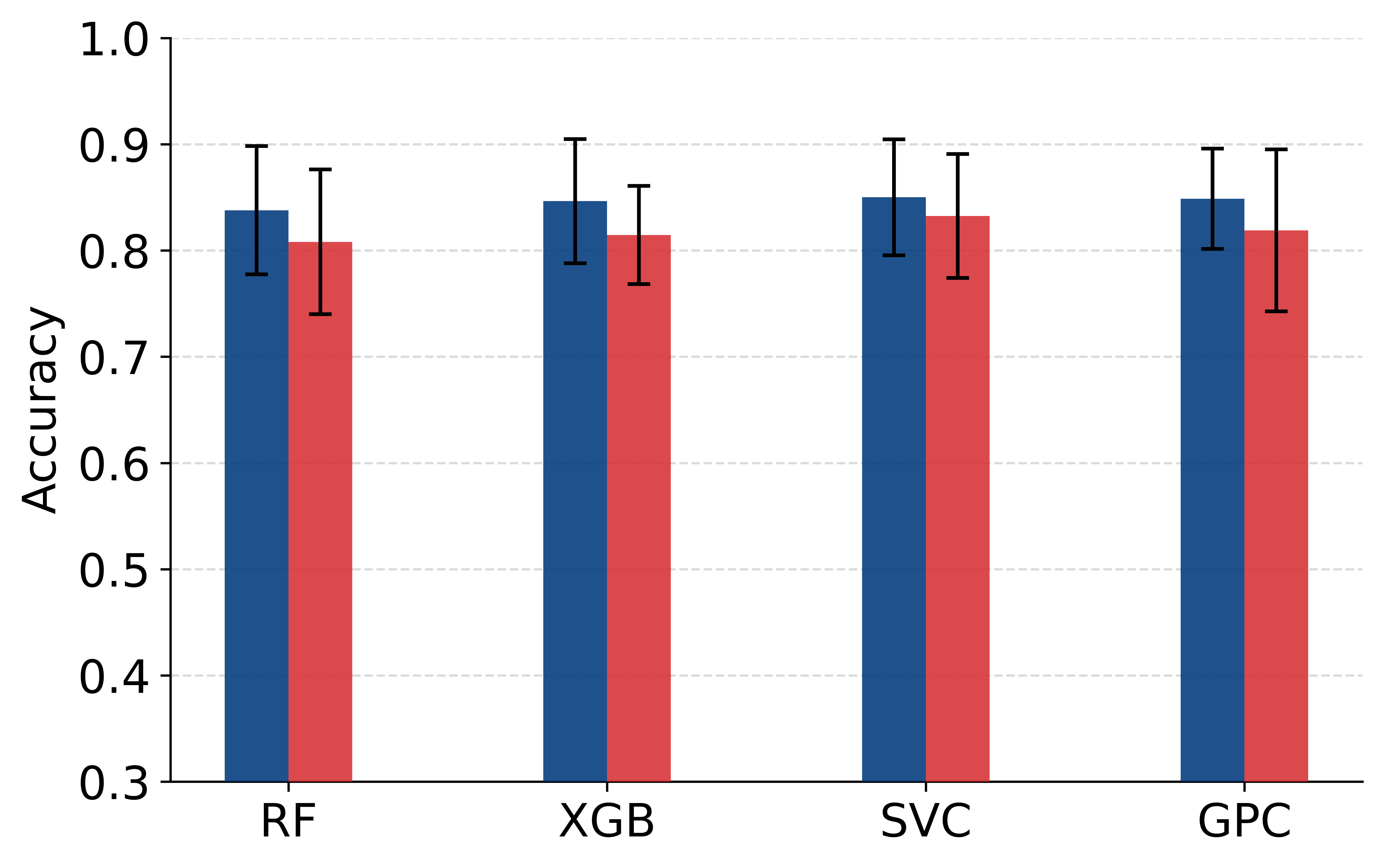}
        \subcaption{}
        \label{fig:5a}
    \end{minipage}
    \hfill
    \begin{minipage}[b]{0.49\linewidth}
        \centering
        \includegraphics[width=\linewidth]{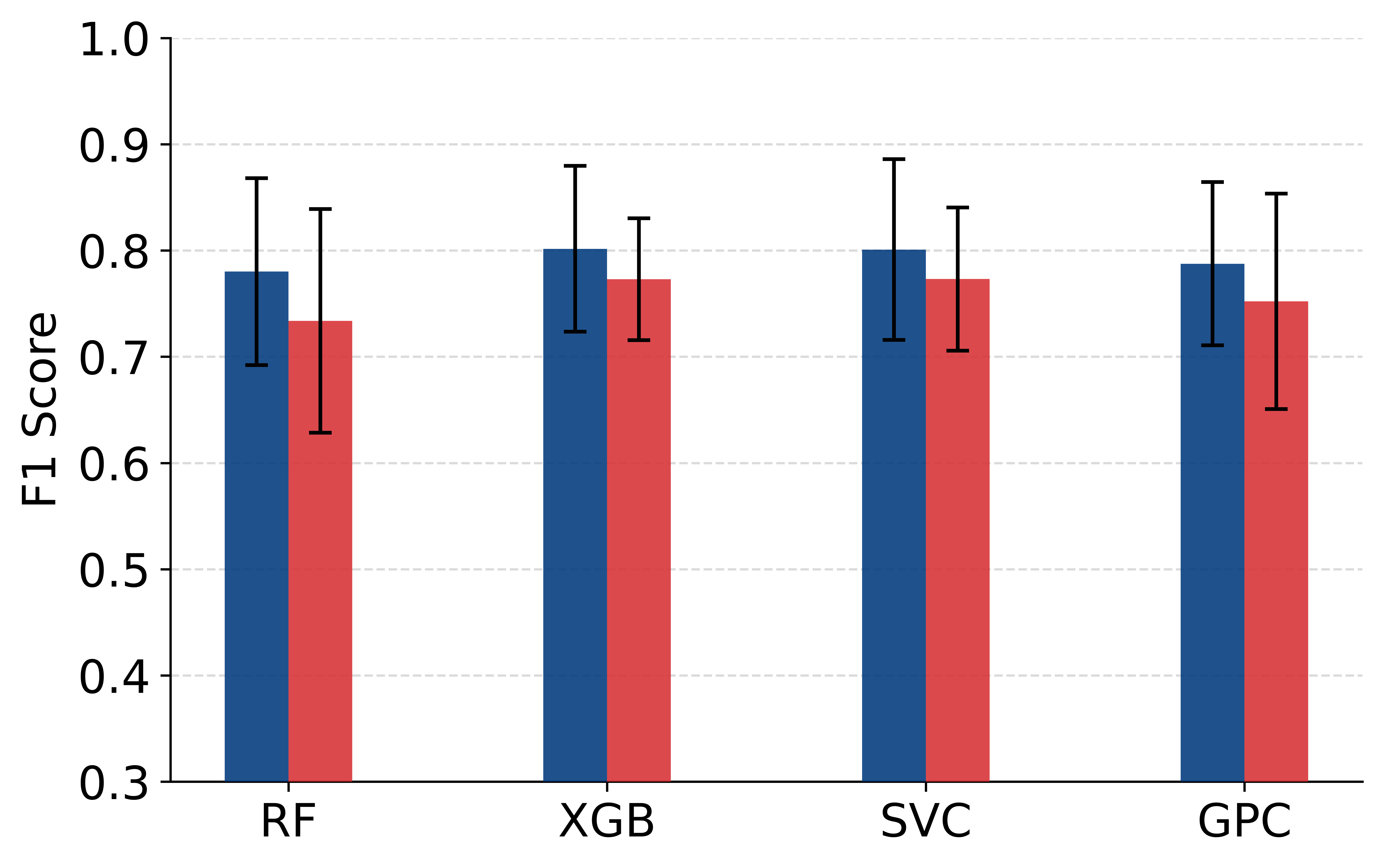}
        \subcaption{}
        \label{fig:5b}
    \end{minipage}

    \par\vspace{0.5ex}

    \begin{minipage}[b]{0.49\linewidth}
        \centering
        \includegraphics[width=\linewidth]{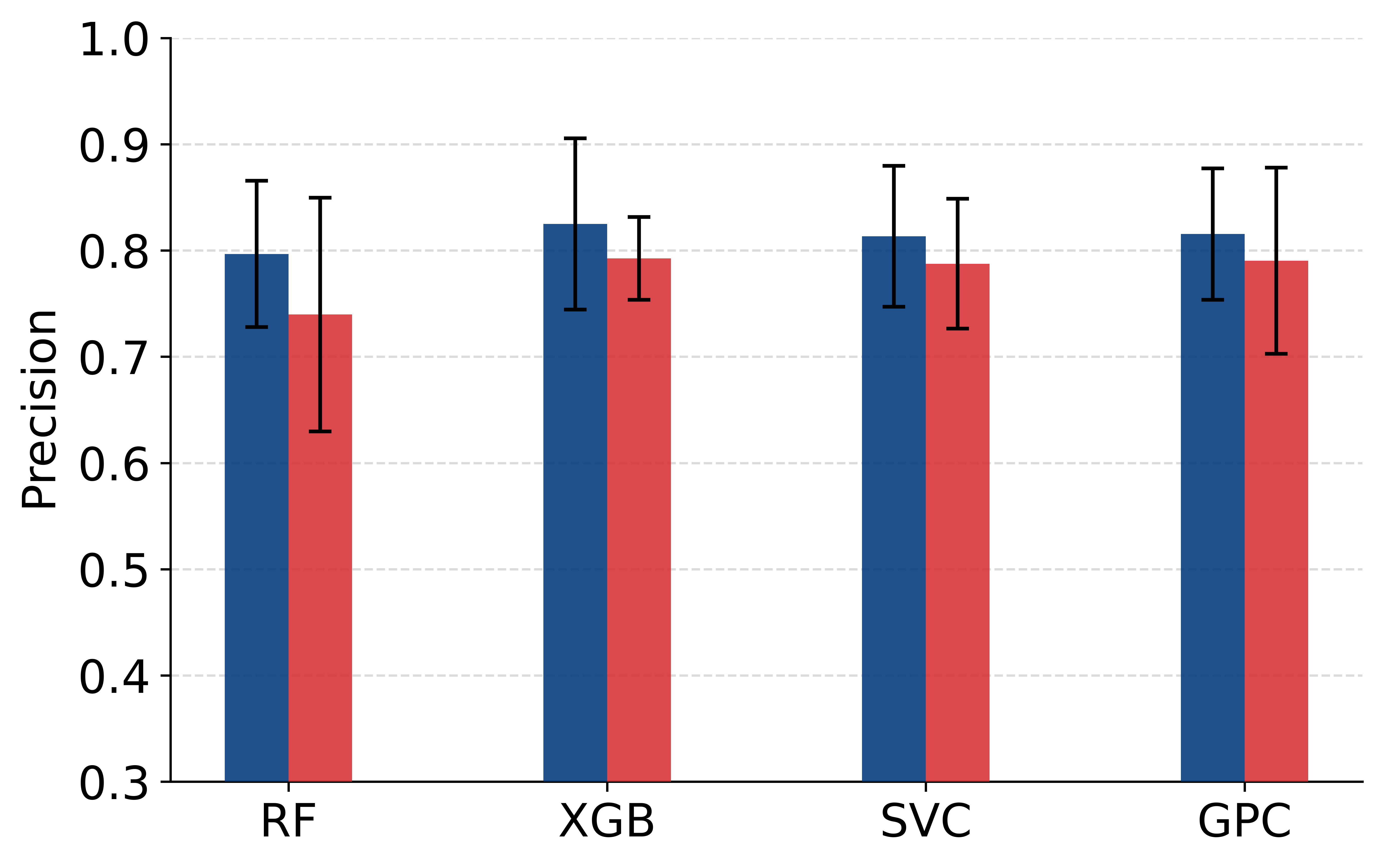}
        \subcaption{}
        \label{fig:5c}
    \end{minipage}
    \hfill
    \begin{minipage}[b]{0.49\linewidth}
        \centering
        \includegraphics[width=\linewidth]{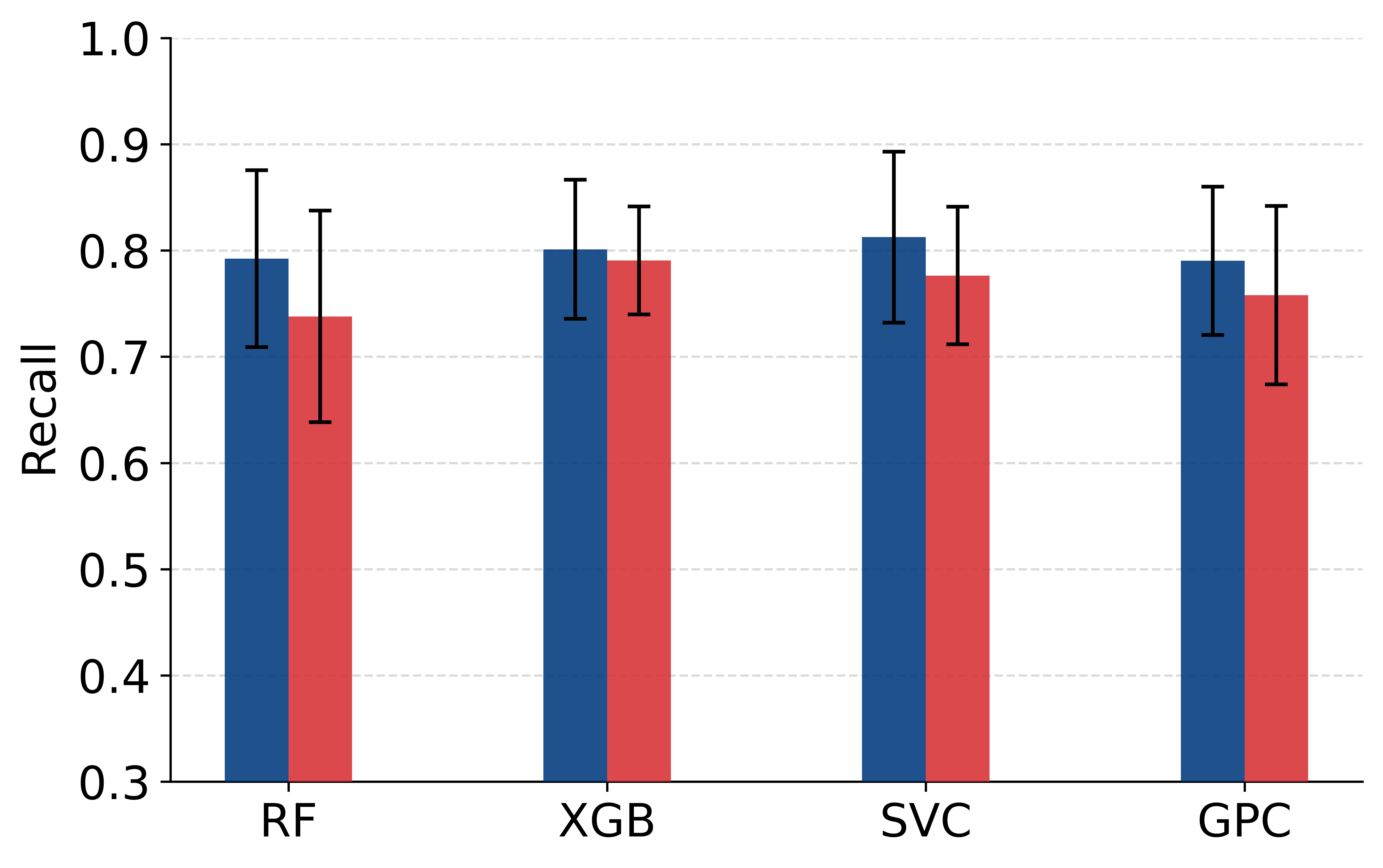}
        \subcaption{}
        \label{fig:5d}
    \end{minipage}

\end{minipage}%
\begin{minipage}[t]{0.16\textwidth}
    \vspace{0pt}
    \centering
    \includegraphics[width=0.65\linewidth]{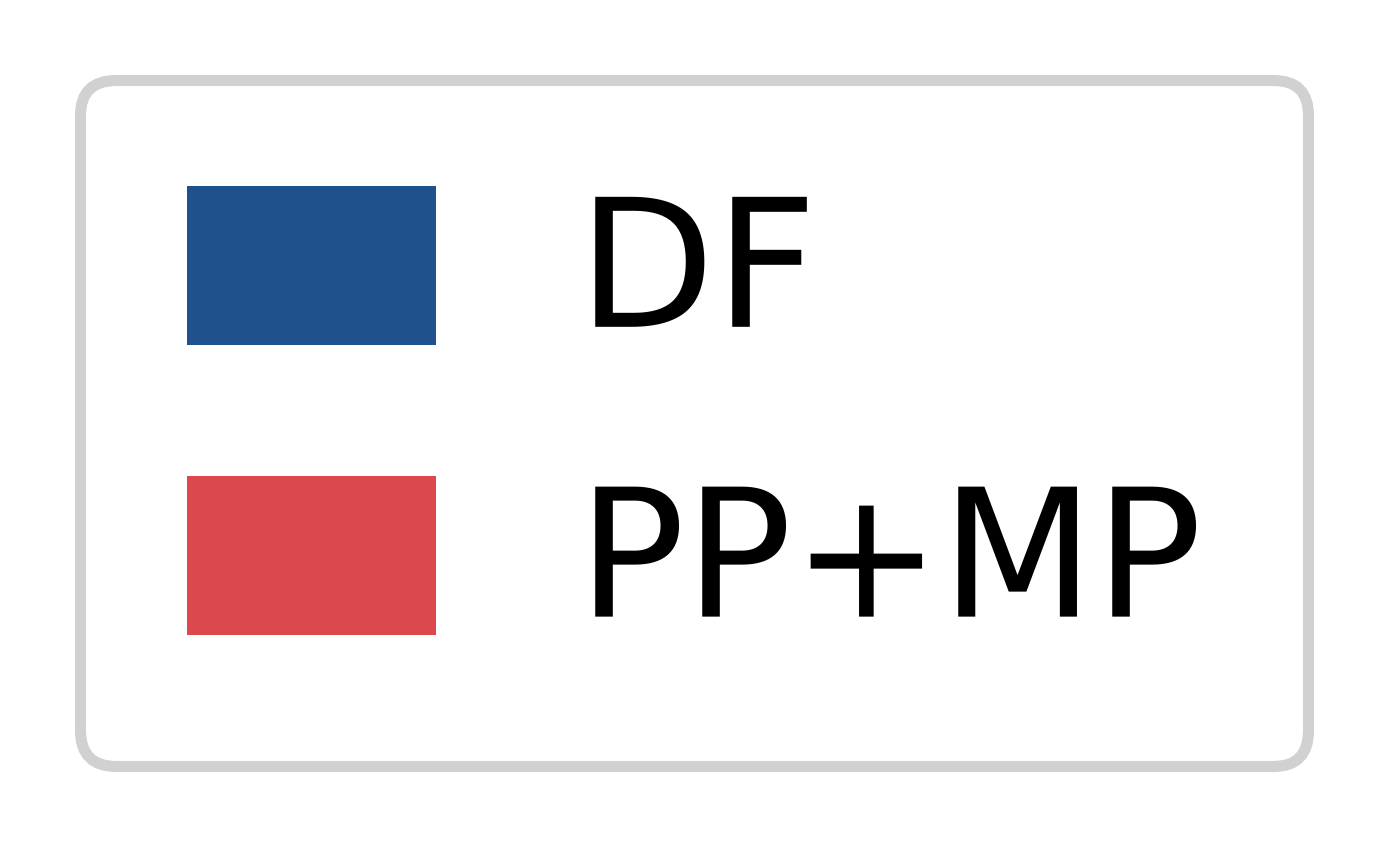}
\end{minipage}

\caption{Performance of ML models for melt pool mode classification evaluated by (a) accuracy, (b) F1 score, (c) precision, and (d) recall using dimensionless features (DF) and process parameters (PP) \& material properties (PP) as inputs for the combined dataset.}
\label{fig:classification_combined}
\end{figure*}

Figure \ref{fig:classification_combined}, Table~\ref{tab:cls_dimless_316L}, and \ref{tab:cls_dimless_nist}
compare dimensionless versus process parameter and material properties
feature representations for classical ML classifiers across the three
datasets based on the performance of the classifiers. Table~\ref{tab:cls_tfm_316L}--\ref{tab:cls_tfm_combined}
then compare classical ML models against TFMs using
the dimensionless representation only. The classification is a three-class
problem (conduction, transition, and keyhole), and metrics reported are
accuracy, macro recall, weighted F1, and macro precision from five-fold
grouped CV.

\begin{table*}[!htb]
\caption{Performance of classical ML models for melt pool mode classification on the 316L dataset using two different input feature sets.}
\label{tab:cls_dimless_316L}
\small
\begin{tabular*}{\hsize}{@{\extracolsep{\fill}}lcccccccc@{}}
\toprule
& \multicolumn{2}{c}{Accuracy}
& \multicolumn{2}{c}{Recall}
& \multicolumn{2}{c}{F1}
& \multicolumn{2}{c}{Precision} \\
\cmidrule(lr){2-3}\cmidrule(lr){4-5}\cmidrule(lr){6-7}\cmidrule(lr){8-9}
Model & DF & PP & DF & PP & DF & PP & DF & PP \\
\midrule
RF  & 0.800 & 0.781 & 0.730 & 0.700 & 0.730 & 0.692 & 0.750 & 0.720 \\
XGB & 0.814 & 0.795 & 0.743 & 0.723 & 0.738 & 0.715 & 0.750 & 0.743 \\
SVC & 0.810 & 0.752 & 0.735 & 0.649 & 0.735 & 0.649 & 0.744 & 0.755 \\
GPC & 0.795 & 0.770 & 0.714 & 0.698 & 0.715 & 0.684 & 0.723 & 0.719 \\
\hline
\end{tabular*}

\smallskip
{\small DF = Dimensionless Features; PP = Process Parameters.}
\end{table*}
 
\begin{table*}[!htb]
\caption{Performance of classical ML models for melt pool mode classification on the IN625+IN718 dataset using two different input feature sets.}
\label{tab:cls_dimless_nist}
\small
\begin{tabular*}{\hsize}{@{\extracolsep{\fill}}lcccccccc@{}}
\toprule
& \multicolumn{2}{c}{Accuracy}
& \multicolumn{2}{c}{Recall}
& \multicolumn{2}{c}{F1}
& \multicolumn{2}{c}{Precision} \\
\cmidrule(lr){2-3}\cmidrule(lr){4-5}\cmidrule(lr){6-7}\cmidrule(lr){8-9}
Model & DF & PP+MP & DF & PP+MP & DF & PP+MP & DF & PP+MP \\
\midrule
RF  & 0.900 & 0.890 & 0.870 & 0.840 & 0.860 & 0.830 & 0.870 & 0.851 \\
XGB & 0.884 & 0.837 & 0.808 & 0.717 & 0.809 & 0.693 & 0.864 & 0.700 \\
SVC & 0.884 & 0.852 & 0.882 & 0.860 & 0.844 & 0.809 & 0.845 & 0.812 \\
GPC & 0.940 & 0.909 & 0.943 & 0.814 & 0.928 & 0.808 & 0.921 & 0.812 \\
\hline
\end{tabular*}

\smallskip
{\small DF = Dimensionless Features; PP+MP = Process Parameters and Material Properties.}
\end{table*}

For the 316L dataset, dimensionless inputs provide consistent but
moderate improvements in classification accuracy. RF accuracy increases
from 0.781 (process parameter and material properties) to 0.800
(dimensionless), XGB from 0.795 to 0.814, and SVC from 0.752 to 0.810.
GPC also improves marginally (0.770 to 0.795). Recall and F1 follow
similar trends, with dimensionless inputs uniformly outperforming
process parameter and material properties inputs for all classical
classifiers. This finding is meaningful because the 316L classification
task is single-material, where one might expect process parameter and
material properties inputs to be competitive. The improvement under
dimensionless inputs suggests that $\Delta H/h_\mathrm{s}$ and $Pe$
provide better discrimination between conduction, transition, and
keyhole regimes than dimensional process parameters, consistent with the
physical interpretation that the aspect ratio $d/w$ – the label
criterion – is more directly related to normalized energy input than
to individual process parameters.
 
For the IN625 + IN718 dataset, the dimensionless advantage is
stronger and more consistent. GPC improves sharply: accuracy
from 0.909 to 0.940, recall from 0.814 to 0.943, and F1 from 0.808 to
0.928. The accuracy of RF improves from 0.890 to 0.900, and SVC from 0.852 to
0.884. The largest gains are in recall and F1 for GPC and SVC,
indicating that the dimensionless representation particularly helps in
correctly identifying the transition and keyhole regimes – the classes
most susceptible to misclassification when material identity is not
properly encoded in the feature space. Under process parameter and
material properties inputs, a single scan speed or power value can
correspond to different regimes depending on the alloy; the
dimensionless parameters remove this ambiguity.
 
In the Combined dataset, dimensionless inputs paired with one-hot
material encoding consistently outperform process parameter and material
properties inputs paired with material encoding across all four
classical classifiers. Accuracy improvements range from 3 to 4
percentage points, and F1 improvements are of similar magnitude. SVC
achieves the highest accuracy among classical models for the Combined
dataset (0.850 dimensionless), followed closely by GPC (0.849) and XGB
(0.846). The consistent improvement across all metrics and classifiers
confirms that the dimensionless representation provides a better
separation of the process-regime manifold than process parameter and
material properties inputs, even when material identity is explicitly
provided to both representations through one-hot encoding.

\begin{figure*}[!htb]
    \centering
    \includegraphics[width=.7\linewidth]{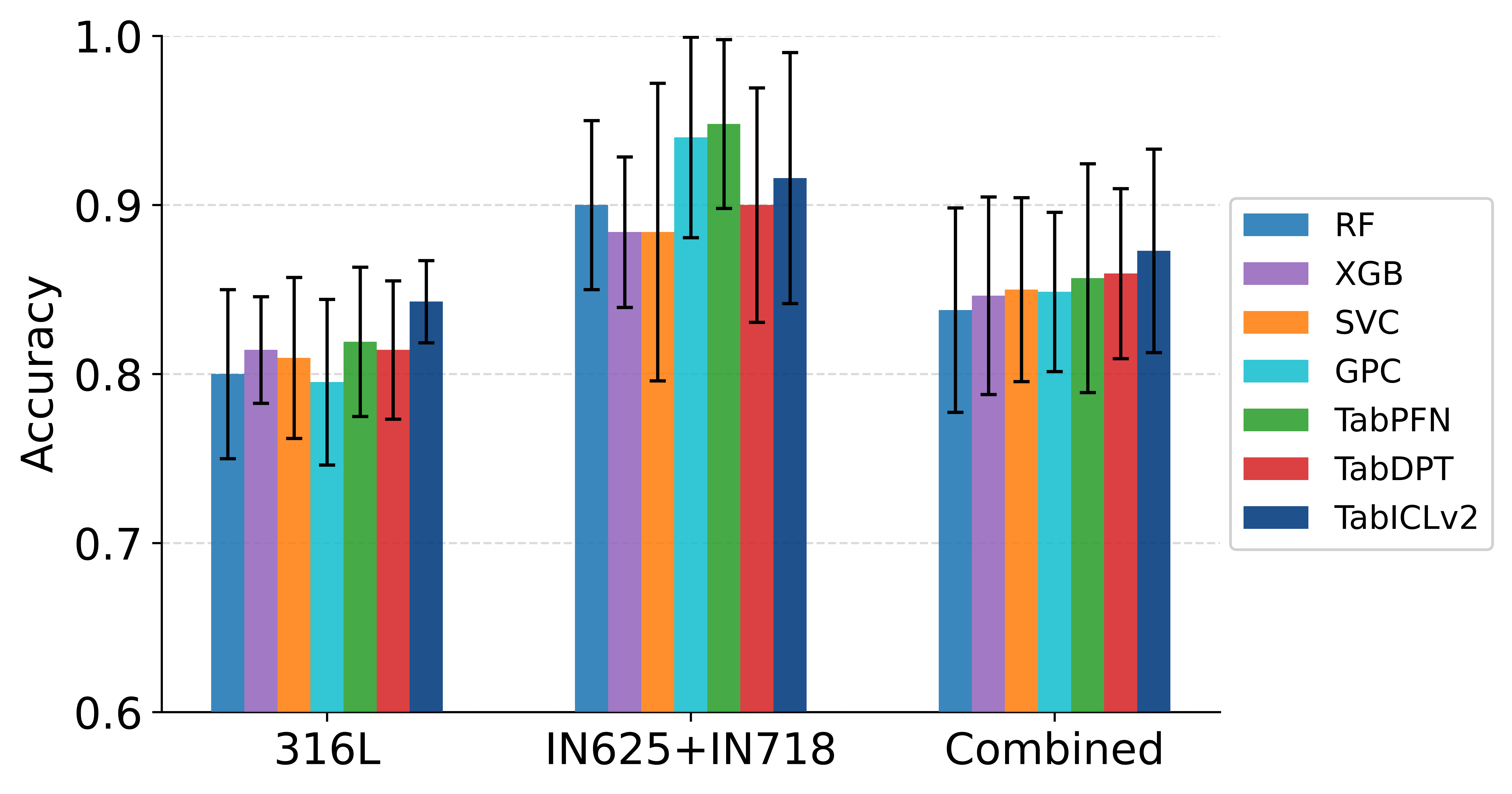}
    \caption{Performance comparison of classical ML models with TFMs for classifying melt pool mode, evaluated by accuracy.}
    \label{fig:Classification accuracy}
\end{figure*}

\begin{table}[!htb]
\caption{Performance of classical ML models and TFMs for melt pool mode classification on the 316L dataset using a dimensionless input feature set.}
\label{tab:cls_tfm_316L}
\small
\begin{tabular*}{\hsize}{@{\extracolsep{\fill}}lccc@{}}
\toprule
Model & Recall & F1 & Precision \\
\midrule
RF       & 0.730 & 0.730 & 0.750 \\
XGB      & 0.743 & 0.738 & 0.750 \\
SVC      & 0.735 & 0.735 & 0.744 \\
GPC      & 0.714 & 0.715 & 0.723 \\
TabPFN   & 0.758 & 0.749 & 0.748 \\
TabDPT   & 0.742 & 0.735 & 0.736 \\
TabICLv2 & 0.762 & 0.764 & 0.786 \\
\hline
\end{tabular*}
\end{table}

\begin{table}[!htb]
\caption{Performance of classical ML models and TFMs for melt pool mode classification on the IN625+IN718 dataset using a dimensionless input feature set.}
\label{tab:cls_tfm_nist}
\small
\begin{tabular*}{\hsize}{@{\extracolsep{\fill}}lccc@{}}
\toprule
Model & Recall & F1 & Precision \\
\midrule
RF       & 0.870 & 0.860 & 0.870 \\
XGB      & 0.808 & 0.809 & 0.864 \\
SVC      & 0.882 & 0.844 & 0.845 \\
GPC      & 0.943 & 0.928 & 0.921 \\
TabPFN   & 0.953 & 0.938 & 0.931 \\
TabDPT   & 0.882 & 0.865 & 0.890 \\
TabICLv2 & 0.915 & 0.902 & 0.904 \\
\hline
\end{tabular*}
\end{table}

\begin{table}[!htb]
\caption{Performance of classical ML models and TFMs for melt pool mode classification on the Combined dataset using a dimensionless input feature set.}
\label{tab:cls_tfm_combined}
\small
\begin{tabular*}{\hsize}{@{\extracolsep{\fill}}lccc@{}}
\toprule
Model & Recall & F1 & Precision \\
\midrule
RF     & 0.792 & 0.780 & 0.797 \\
XGB    & 0.801 & 0.802 & 0.825 \\
SVC    & 0.813 & 0.801 & 0.813 \\
GPC    & 0.790 & 0.788 & 0.815 \\
TabPFN & 0.813 & 0.803 & 0.819 \\
TabDPT & 0.812 & 0.801 & 0.819 \\
TabICLv2     & 0.835 & 0.828 & 0.840 \\
\hline
\end{tabular*}
\end{table}

Comparing TFMs against ML models within the dimensionless representation reveals dataset-dependent patterns. For the 316L dataset, TabICLv2 achieves the highest accuracy (0.843) and F1 (0.764) among all evaluated models, outperforming the best classical ML model, XGBoost (XGB), which achieves an accuracy of 0.814 and an F1-score of 0.738. TabPFN and TabDPT also demonstrate competitive performance, achieving accuracies of 0.819 and 0.814, respectively, comparable to those of classical ensemble methods. However, these improvements are modest, suggesting that the knowledge acquired through TFM pretraining provides only a limited advantage over well-tuned classical classifiers for relatively small, single-material classification tasks.
 
The IN625+IN718 dataset reveals a more pronounced advantage for TFMs over classical classifiers. TabPFN achieves the highest accuracy (0.948) and recall (0.953) across all evaluated models and datasets, while its F1-score (0.938) exceeds that of the best classical classifier, GPC (0.928). TabICLv2 also outperforms RF and XGB. Notably, GPC is the strongest classical classifier on
IN625+IN718, achieving an accuracy of 0.940, potentially reflecting the ability of kernel-based probabilistic modeling to capture nonlinear decision boundaries. However, TabPFN outperforms even GPC, suggesting that knowledge acquired through TFM pretraining can provide additional predictive benefits in this multi-alloy classification problem.
 
For the Combined dataset, TabICLv2 again achieves the highest
performance among all models (accuracy 0.873, recall 0.835, F1-score 0.828, and
precision 0.840), followed by TabDPT (accuracy 0.859) and TabPFN
(accuracy 0.857). Among classical models, SVC performs best (accuracy 0.85). The
TFM advantage in the Combined dataset is modest in absolute terms, but all three TFMs achieve higher accuracy than the best classical classifier. These findings suggest that TFMs can provide incremental improvements in multi-material melting-regime classification when physics-informed dimensionless inputs are used.

\subsection{Summary of Key Findings}

The results across the three datasets and two prediction tasks support
two principal conclusions.

First, physics-informed dimensionless
representation ($\Delta H/h_\mathrm{s}$, $Pe$) provides comparable or superior performance in predicting melt pool depth and width compared to dimensional features used as inputs in most evaluated conditions. The improvements are most pronounced for melt pool depth prediction – where thermal penetration physics are directly
encoded in $\Delta H/h_\mathrm{s}$ – and for multi-material datasets
(IN625+IN718 and combined). These findings suggest that dimensionless representations, which characterize energy input and heat transport while accounting for material-specific thermophysical properties, can improve predictive performance across different alloys. However, for melt pool width prediction using the single-material 316L dataset, conventional inputs perform comparably or marginally better, indicating that the benefits of dimensionless representations depend on the prediction target and dataset characteristics. 

Second, tabular foundation models generally match or outperform classical machine learning models, with the most pronounced improvements observed on the multi-material datasets. For the single-material 316L dataset, the performance differences are relatively small, and classical models such as GPR, SVR, and XGB remain competitive. 

Based on the observations, combining physics-informed dimensionless representations with TFMs yields particularly strong results for melt pool depth prediction and melting-mode classification. These findings demonstrate the potential of integrating physics-informed feature representations with pretrained TFMs, while highlighting that their relative advantages depend on the prediction task and material composition of the dataset.

\section{Conclusion}
\label{Conclusion}

This study developed and evaluated a systematic data-driven framework to predict melt pool geometry and classify the prevailing melting mode in laser powder bed fusion (LPBF) using two distinct feature representations and seven machine learning models. The evaluations employed process-condition-grouped cross-validation to ensure rigorous, leakage-free assessment of generalization.

Across the evaluated conditions, dimensionless representations improve generalization, particularly across materials and for depth prediction. Physics-informed dimensionless inputs outperformed raw process parameters and material properties in most of the evaluated conditions. The advantage was most pronounced for melt pool depth prediction – where thermal penetration is directly encoded in the normalized enthalpy – and for multi-material datasets. On the combined dataset, dimensionless inputs improved $R^2$ by up to 4.9\% (TabICLv2: 0.907 to 0.956 for depth prediction), and reduced NRMSE by 10\%-15\% across models. On the Combined dataset, the dimensionless representation delivered consistent improvements for all models and both targets, confirming that dimensionless enthalpy and Peclet number act as material-agnostic descriptors that collapse multi-material data onto a common process-property manifold. The benefits of this representation were complemented by the predictive capability of tabular foundation models. Among all model-representation combinations evaluated, TabPFN achieved the highest $R^2$ for depth prediction on the Combined dataset ($R^2$ 0.961, NRMSE 0.043) – the best result across all experiments – and the highest classification accuracy on the IN625+IN718 dataset (accuracy 0.948, recall 0.953). TabICLv2 achieved the strongest classification performance on the 316L and Combined datasets. The advantage of tabular foundation models over classical methods was clearest on multi-material datasets and smallest on the single-material 316L dataset, where classical models such as GPR ($R^2$ = 0.930 for depth) remained competitive. Critically, tabular foundation models produced more uniformly distributed prediction errors than classical kernel methods. Combining dimensionless inputs with pretrained tabular models offers the strongest and most consistent framework. Across regression and classification tasks, pairing physics-informed dimensionless representations with tabular foundation models yielded the best generalization in most conditions. This combination is particularly well suited to LPBF process qualification scenarios where experimental data are scarce, materials are diverse, and process conditions span wide ranges of energy input and scanning geometry. The results provide quantitative ML-level evidence for a hypothesis previously supported only by analytical scaling arguments -- that dimensionless parameters enable cross-material knowledge transfer in LPBF.

Despite the encouraging findings, several limitations of this study should be acknowledged. The normalized enthalpy was computed using a constant laser absorptivity for all materials and process conditions. In practice, absorptivity varies with surface state, powder characteristics, and melt pool regime – particularly in the keyhole mode, where multiple reflections significantly increase effective energy absorption. The dataset was also limited to two alloy families – 316L stainless steel and IN625 \& IN718 nickel superalloys. The generalizability of the findings to other alloy systems with substantially different thermophysical properties, reflectivity, and solidification behavior is yet to be explored. In addition, the GPR configuration used a single isotropic RBF component, which constrains all standardized input dimensions to a common length scale. This restriction may limit its flexibility for the raw representation, whose predictors describe distinct physical effects.

This study's findings open several productive directions for future research. Validating the dimensionless scaling framework on aluminum alloys, titanium alloys, and copper – materials with fundamentally different thermal conductivity and melting behavior – would significantly strengthen the cross-material generalization claim. Including powder layer thickness and hatch spacing as additional raw inputs, and investigating whether further dimensionless groups (e.g., a Stefan number or normalized scanning speed) improve predictive performance, are natural extensions. Replacing the constant absorptivity assumption with a dynamic absorptivity model – for example, using the empirical relations of Trapp et al. (2017) \cite{TRAPP2017341} or the thermophysical model of Gan et al. (2021) \cite{gan2021universal} – would improve the physical fidelity of the dimensionless representation and is expected to particularly benefit predictions in the keyhole regime. Integrating absorptivity as a derived input feature, computed from process parameters and melt pool geometry feedback, would create a tighter coupling between physics and data. Applying the dimensionless input framework to predict porosity, keyhole stability index, and primary dendritic arm spacing – targets with direct structural consequence – would extend the practical relevance of the framework beyond geometric proxies. Combining normalized enthalpy and Peclet number with microstructure outputs has strong physical motivation from the solidification literature and represents a natural next step.

A second direction is to examine whether adaptation of pretrained TFMs to LPBF data improves performance beyond in-context learning alone. Investigating whether lightweight fine-tuning on LPBF-specific data (e.g., via LoRA-style adapters or continued pretraining on synthetic thermal datasets) improves performance, particularly for depth prediction on small single-material datasets, is an important open question for the emerging field of scientific tabular foundation models. The dimensionless input framework is well suited to integration with real-time process monitoring systems, where pyrometer or photodiode data can be used to estimate melt pool regime and geometry on the fly. Coupling the trained models with digital twin platforms – where online sensor data updates the model predictions as a build progresses – would enable adaptive process control that maintains dimensional accuracy and avoids defect conditions across varying build geometries and scan strategies.

This study demonstrates that physics-informed dimensionless representations and pretrained tabular foundation models, used in combination, constitute a principled and practically effective framework for data-efficient melt pool geometry prediction in LPBF. The framework is material-agnostic, requires only independently measurable inputs, and generalizes across alloy systems without material-specific retraining. These properties make it a strong candidate for integration into next-generation smart manufacturing workflows where process qualification speed, cross-material transferability, and data efficiency are paramount.

\bibliographystyle{plainnat}
\bibliography{my_collection}

\end{document}